\documentclass{article}

\usepackage{lmodern}
\usepackage[T1]{fontenc}
\usepackage{kpfonts}

\usepackage{graphicx}%
\usepackage{multirow}%
\usepackage{amsmath,amssymb,amsfonts}%
\usepackage{amsthm}%
\usepackage{mathrsfs}%
\usepackage[title]{appendix}%
\usepackage{xcolor}%
\usepackage{textcomp}%
\usepackage{manyfoot}%
\usepackage{algorithm}%
\usepackage{algorithmicx}%
\usepackage{algpseudocode}%
\usepackage{listings}%
\theoremstyle{plain}

\theoremstyle{remark}

\theoremstyle{definition}

\usepackage{arydshln}
\usepackage[numbers]{natbib}
\usepackage{algorithm}
\usepackage{makecell} 

\usepackage{longtable}
\usepackage{lscape}
\usepackage{adjustbox}
\usepackage{float} 
\usepackage{pdflscape}
\usepackage{amsmath}
\usepackage{pifont}  
\usepackage{multicol}

\usepackage{array}
\usepackage{colortbl}

\usepackage{booktabs}
\usepackage{tabularx}
\usepackage{geometry}
\usepackage{multirow}
\usepackage{lettrine}
\newcommand{\cmark}{\textcolor{blue}{\ding{51}}}  
\newcommand{\xmark}{\textcolor{red}{\ding{55}}}  

\usepackage{graphicx}
\usepackage{multirow}
\usepackage{float}
\usepackage{array}
\usepackage{colortbl}
\usepackage{hyperref}
\hypersetup{
    colorlinks=true,
    linkcolor=blue,
    citecolor=blue,
    urlcolor=blue
}
\usepackage{algorithm}
\usepackage{algpseudocode}
\usepackage{listings}
\usepackage{longtable}
\usepackage{adjustbox}
\usepackage{lscape}
\usepackage{pifont}
\usepackage{tabularx}
\usepackage{lettrine} 

\usepackage{authblk}

\title{Multimodal Generative AI with Autoregressive LLMs for Human Motion Understanding and Generation: A Way Forward}

\author[1]{Muhammad Islam\thanks{muhammad.islam1@my.jcu.edu.au}}
\author[1]{Tao Huang}
\author[1]{Euijoon Ahn}
\author[2]{Usman Naseem}

\affil[1]{College of Science and Engineering, James Cook University, Cairns QLD 4878, Australia}
\affil[2]{School of Computing, Macquarie University, Australia}

\date{}  

\begin{document}

\maketitle
\begin{abstract}

This paper presents an in-depth survey on the use of multimodal Generative Artificial Intelligence (GenAI) and autoregressive Large Language Models (LLMs) for human motion understanding and generation, offering insights into emerging methods, architectures, and their potential to advance realistic and versatile motion synthesis. Focusing exclusively on text and motion modalities, this research investigates how textual descriptions can guide the generation of complex, human-like motion sequences. The paper explores various generative approaches, including autoregressive models, diffusion models, Generative Adversarial Networks (GANs), Variational Autoencoders (VAEs), and transformer-based models, by analyzing their strengths and limitations in terms of motion quality, computational efficiency, and adaptability. It highlights recent advances in text-conditioned motion generation, where textual inputs are used to control and refine motion outputs with greater precision. The integration of LLMs further enhances these models by enabling semantic alignment between instructions and motion, improving coherence and contextual relevance. This systematic survey underscores the transformative potential of text-to-motion GenAI and LLM architectures in applications such as healthcare, humanoids, gaming, animation, and assistive technologies, while addressing ongoing challenges in generating efficient and realistic human motion.

\end{abstract}

\vspace{1em}

\maketitle
\section{Introduction}
 \lettrine{H}{\textbf{uman}} motion generation using multimodal generative AI particularly through the integration of autoregressive Large Language Models (\textcolor{blue}{LLMs}) has emerged as a promising research frontier, driving advancements toward the broader goal of Artificial General Intelligence (\textcolor{blue}{AGI})~\cite{1}. Recent examples, such as SORA \cite{2}, GPT-4V \cite{3}, and Dall-E \cite{4} showcase advancements in generative domains conditioned on signals to create motion, video, and images. Human motion generation \cite{4a}, in particular, has emerged as a crucial field, driven by these Generative Artificial Intelligence \textcolor{blue}{(GenAI)} advancements and the increasing demand for realistic, human-like movements in applications like film production, video games, robotics, healthcare, autonomous vehicle perception, and virtual training environments. This survey focuses on the intersection of text-conditioned motion generation using LLM and diffusion models representing significant breakthroughs in enabling machines to produce human motions based on natural language descriptions.
 The rise of generative models such as Generative Adversarial Networks (GANs) \cite{5}, Variational Autoencoders (VAEs) \cite{6}, Normalizing Flows \cite{7}, Autoregressive \textcolor{blue}{(AR)} \cite{8} and Diffusion Models \cite{9} \cite{10} has been central to improving the quality and diversity of generated motions. These models, while excelling in image generation and other domains, have started making strides in the challenging task of generating human motions that are plausible, natural, controllable, and contextually appropriate based on text prompts.
 
 This survey paper aims to furnish a comprehensive and insightful review of human motion generation by reviewing more than 250 research articles from prestigious and top-rated journals and conferences, which include CVPR, ICCV, ECCV, SigIR, NIPS, TIPS, TPAMI, and 3DV over recent years.  We also critically reviewed the methodologies and backbone architectures,  highlighting the technical challenges posed by human motion’s inherent complexity and how recent innovations in LLMs like GPT-4 \cite{11}, Llama-3 \cite{12}, BERT \cite{13}, and T5 \cite{14} are being leveraged to condition motion understanding based on text. By integrating diffusion models \cite{4a}, GANs \cite{16}, and VAEs \cite{17} variants into text-to-motion pipelines, we explore how \textcolor{blue}{AI} can move closer to mimicking human-like movements by creating digital human actions and how to achieve generalized and robust motions with low computational cost. The work also keeps track of and highlights the progress of motion understanding and generation based on condition signals, trends, and challenges and how a unified model helps achieve the required goal. This work will serve as a valuable resource for researchers and practitioners interested in advancing the state-of-the-art in motion generation through multimodal \textcolor{blue}{AI} approaches. 
 The need for this survey is crucial to understand the human motion generation from low level to current multimodal practices using \textcolor{blue}{GenAI} and \textcolor{blue}{Multimodal LLM}. Some of the recent survey papers published are broader to video \cite{18} and images \cite{19} with motions or based on downstream subtopics of human motion like prediction \cite{20}. Some surveys are just for generative methodologies general understanding like GAN, VAE, and diffusion \cite{6}. Similarly, one of the surveys covers all the general perspectives and condition-based motion generation excluding multimodal LLMs and transformers \cite{21}. Recently, a survey for multimodal LLM and multimodal generative AI includes text, audio, videos, and images but does not cater explicitly for human motion generation \cite{22}. Our work is the first, and unique attempt that covers all the areas of Human Motion Understanding and Generation (\textbf{HMUG}).  Table~\ref{tab:1} showcases a comparative analysis between our work and other surveys.

\begin{figure*}
    \centering
    \includegraphics[width=1\linewidth]{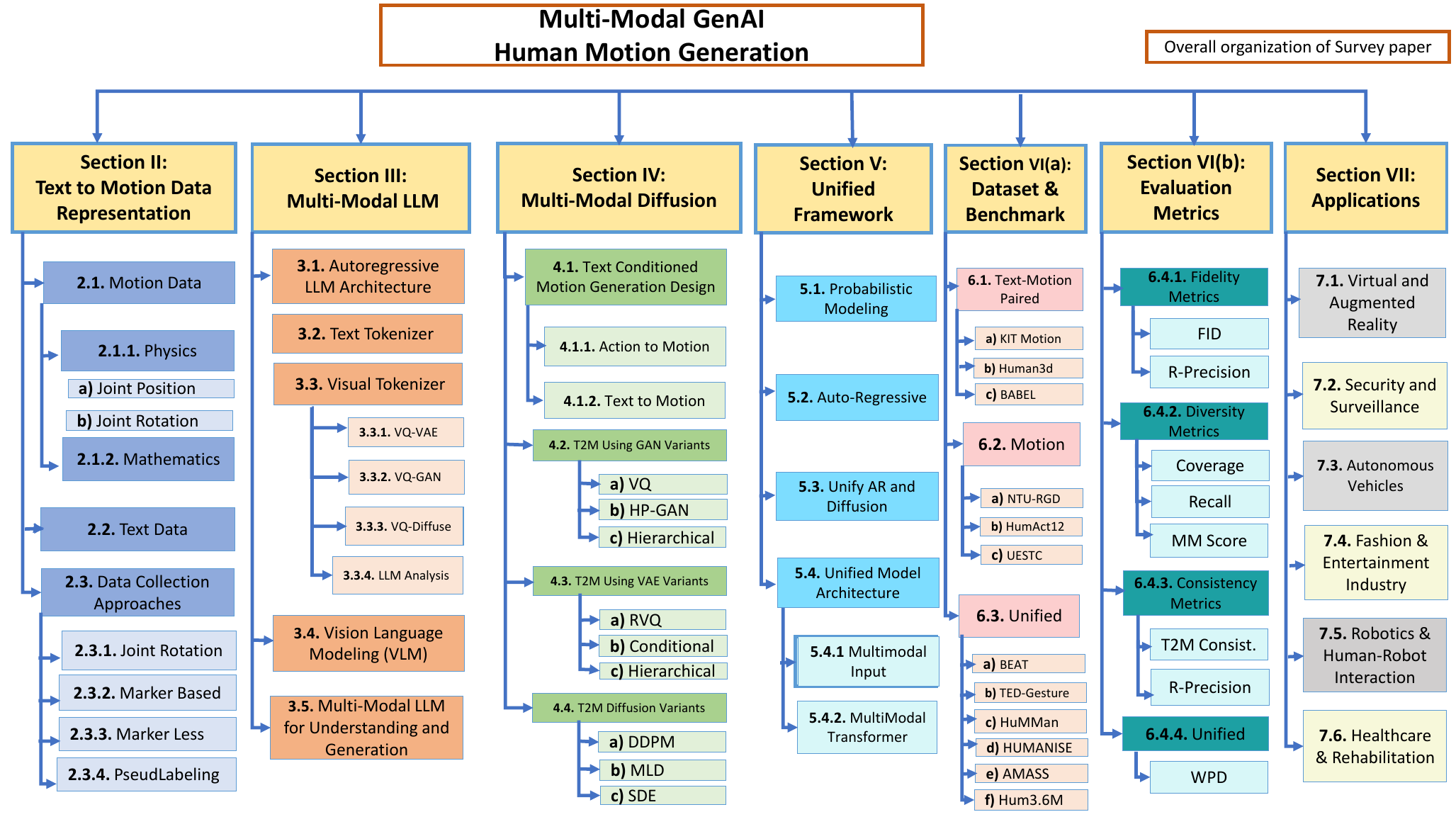}
    \caption{The overall flow diagram illustrating variants of multimodal generative models, multimodal LLM architectures, and unified frameworks along with their backbone architectures. }
   \label{FIG:1}
\end{figure*}

\textbf{Scope:} 
 The scope of this survey focuses on the synthesis of human motion based on multimodal text-conditioned models, particularly using \textcolor{blue}{LLMs} with \textcolor{blue}{GenAI}. While it includes simpler text-to-motion models that inform LLM-based research, the survey excludes other conditioning factors, such as audio, scene context, and image and video inputs, thereby narrowing its focus to textual descriptions that drive more generalized and robust motion generation through the integration of both models. 
 In terms of motion generation, the survey emphasizes both 2D and 3D representations, incorporating core physics and mathematical principles, such as kinematics, dynamics, and transformations, to simulate realistic human motion. It also highlights the role of keypoints and joint rotation sequences—details often overlooked in previous literature. The survey provides an in-depth analysis of \textcolor{blue}{GenAI} methodologies, covering architectural variants and backbones to guide researchers in selecting suitable models for their needs.
 This survey is intended to provide a foundational overview for comprehending multimodal human motion generation, discussing current challenges and opportunities, pipelines, evaluation metrics, and applications in both \textcolor{blue}{real and digital} worlds, and exploring potential future architectural approaches in this evolving field. Figure~\ref{FIG:1} illustrates the overall flow of the paper with sections and subsections. 
\section{Preliminary of Human Motion Data  }
We begin our discussion by introducing how human pose and motion data are represented and processed. This includes physical and mathematical formulations, as well as text-based conditioning and motion data collection techniques, which are essential for training generative models.
\subsection{Pose Data Representation}

In the context of 3D human motion, several popular representations are commonly employed, each offering unique advantages depending on the specific task  \cite{22a}. These include point clouds, which model the human body using individual points distributed in 3D space; voxels, which capture volumetric information by dividing space into a grid of cubes; meshes, which use interconnected vertices and edges to represent surface geometry; and depth maps, which encode distance information from a specific viewpoint.
Neural fields and hybrid representations combine these traditional approaches with neural networks to model continuous 3D geometry and appearance, providing more flexible and detailed descriptions of human motion. These representations form the basis for developing accurate motion generation models, allowing for the simulation of complex human movements with high fidelity. Understanding these representations is crucial for analyzing human motion in various fields, from computer vision to biomechanics.

\subsubsection{Physical Representation}

In motion analysis, physical representation is essential for accurately modeling, predicting, and generating human movements. Two key strategies to address are overall representation of the body structure, utilizing the kinematic chain to understand human motion \cite{23}. The second one is the motion laws \cite{24}, which depend on the body-specific part motion and its relevance to other body parts in a spatial-temporal environment \cite{20}. 
Two further key components of physical representation are joint positions \cite{25, 26} and joint rotations, which together provide a comprehensive understanding of human body motion dynamics. Below, we explore these two concepts and their importance in the field of human motion analysis, prediction, and generation.

\textbf{Joint Positions}
 Refer to the 2D or 3D coordinates of specific key points on the human body, such as knees, elbows, and shoulders. These positions form the skeletal structure and are often used in motion analysis to track and represent body movements over time. By mapping the relative positions of these joints, motion systems can capture how different body parts move through space, making \textcolor{blue}{Joint Positions (JPs)} a robust foundation for motion representation.
This key point-based approach is commonly applied in motion capture systems and has been adopted in various models, such as SkelNet \cite{23}, CHA \cite{27}, MGCN \cite{28}, and AM-GAN \cite{29}. These models rely on joint positions to effectively track and predict human motion in real time. Joint positions are intuitive and provide direct visual data, making them highly interpretable and useful for applications such as animation, robotics, and biomechanics. However, JPs alone may not capture the complete range of motion details, especially when it comes to the angular behavior of body parts. This is where joint rotations play a vital role, often necessitating the conversion of joint positions into joint rotations using techniques like inverse kinematics (IK).
Recent advancements have expanded this keypoint-based representation by introducing body markers \cite{30}, which provide more detailed insights into body shapes and limb twists compared to traditional skeleton-based keypoint \cite{31, 32}. This extended approach enables a richer and more comprehensive analysis of motion, providing a better understanding and interpretation of complex body movements.

\textbf{Joint Rotations} While joint positions emphasize on the spatial layout of the biomechanical structure, joint rotations offer insights into the angular motion of joints, describing how one joint rotates relative to its parent in a hierarchical structure. \textcolor{blue}{Joint Rotations (JR)} are critical for capturing the true complexity of human motion, such as limb twisting or joint bending. These rotations can be parameterized using formats like Euler angles, axis angles, or quaternions, all of which provide different ways to describe joint movements.
In rotation-based models, such as the Skinned Multi-Person Linear (\textcolor{blue}{SMPL}) model \cite{33}, the body is represented through a combination of pose and shape parameters, where the pose parameters (joint rotations) describe the angular relationship between joints. The \textcolor{blue}{SMPL} model uses these parameters to generate a detailed 3D mesh of the human body \cite{34}, including 24 joints that define the skeleton's rotations. The shape parameters, on the other hand, capture individual body dimensions, such as height. This fusion of joint rotations and shape parameters offers a detailed and dynamic representation of human motion, widely utilized in applications such as virtual reality, animation, and biomechanics.
Furthermore, SMPL-X \cite{33, 34}, an extension of the SMPL model, incorporates face and hand movements, offering a more holistic view of body motion. Beyond SMPL, other statistical models, such as GHUM and STAR \cite{35, 36}, have been developed to provide alternative approaches to human body modeling. Incorporating joint rotations enables models to capture dynamic properties, such as velocity and acceleration, thereby enhancing the precision and realism of motion predictions. By integrating both JPs and JRs, motion analysis systems achieve a more complete and nuanced representation of human motion.

\subsubsection{Mathematical Representation}

Mathematical representations play a crucial role in human motion analysis by providing abstract descriptions of human poses, allowing researchers to apply prior knowledge to motion generation and prediction. These representations, derived from motion capture data, map human poses to various mathematical spaces and distributions, simplifying feature extraction for machine learning models. The existing representations are arranged in math encodings such as algebra \cite{37, 38, 39},  differential encoding \cite{28, 40}, graph \cite{40, 41, 42, 43} and motion trajectory \cite{44, 45}. Key approaches include the use of joint positions and joint rotations, each of which provides a unique perspective on motion dynamics. 

\textbf{Joint Positions} are a fundamental aspect of representing human motion mathematically, where the human body is modelled as a set of key points, typically corresponding to anatomical landmarks such as joints (e.g., knees, elbows, and shoulders). These positions, captured in 2D or 3D, provide the spatial coordinates necessary for tracking human movements. Motion capture systems often utilize these \textcolor{blue}{JPs} to describe the movement of different body parts through space, making them a vital tool for real-time applications such as animation, robotics, and biomechanics.
In mathematical models, \textcolor{blue}{JPs} are utilized to create a skeletal structure, enabling algorithms to track motion trajectories over time. These positions are frequently parameterized in terms of algebraic structures, making it easier to capture dependencies between joints. For example, models such as SRNN, MGCN, and MT-GCN \cite{28} \cite{46} \cite{47} rely on \textcolor{blue}{JP} data to generate motion by analyzing how key points change over time.
However, representing motion purely through \textcolor{blue}{JPs} has its limitations, as it cannot capture detailed angular movement and limb orientation. To address these gaps, \textcolor{blue}{JRs} are often used in conjunction with \textcolor{blue}{JPs}, allowing for a more comprehensive and accurate analysis of motion.

\textbf{Joint Rotations} provide a more intricate view of motion by describing the angular movement of body parts relative to each other. This rotation-based representation is crucial for capturing the full complexity of human motion, including twisting, bending, and rotating movements. \textcolor{blue}{JRs} are typically represented using formats such as Euler angles, axis angles, or quaternions \cite{37}, each offering a mathematical approach to model rotational behavior without encountering issues like discontinuities.
One of the key models that utilizes \textcolor{blue}{JRs} is the SMPL model, which represents the human body through a combination of \textcolor{blue}{JRs} (pose parameters) and body dimensions (shape parameters). The SMPL model and its extensions, such as SMPL-X \cite{33}, are widely used for generating detailed 3D meshes that capture both the body’s shape and its motion. This form of representation is critical in domains where realistic modeling is key, including virtual reality and biomechanics.
Mathematically, \textcolor{blue}{JRs} are often encoded into algebraic \cite{37} \cite{38} \cite{39} or differential structures \cite{28} \cite{40} to facilitate analysis. For example, quaternion representation, commonly used in models like QuaterNet \cite{37}, avoids the problem of singularities, making it a stable and efficient way to represent rotations. Additionally, some models use Lie groups \cite{48} to capture the geometry of joint relationships, enabling more sophisticated motion generation.
By combining \textcolor{blue}{JPs} and \textcolor{blue}{JRs}, mathematical models can produce a rich, dynamic representation of human motion that is well-suited for generation and predictive tasks, animation, and real-time motion tracking.
In summary, mathematical representations of \textcolor{blue}{JPs} and rotations form the backbone of modern human motion generation models. Together, they provide a robust framework for understanding and simulating human motion, bridging the gap between abstract pose data and practical applications.

\begin{table}[t]
\caption[Summary of related surveys]{The summary of related surveys. Prior works have primarily focused on specific tasks within human centered motion generation, often lacking a unified multimodal perspective. They also tend to overlook the architectural frameworks, backbone models, and the emerging role of LLMs in motion generation. In contrast, our survey offers a comprehensive overview, covering both human motion understanding and generation, while also categorizing models by their architectures, backbone types, and application domains.}
\label{tab:1}
\centering 
\renewcommand{\arraystretch}{1.3} 
\setlength{\tabcolsep}{3pt} 
\footnotesize
\begin{tabular}{%
>{\arraybackslash}p{3.6cm} 
>{\centering\arraybackslash}p{0.7cm} 
>{\centering\arraybackslash}p{0.7cm} 
>{\centering\arraybackslash}p{0.9cm} 
>{\centering\arraybackslash}p{0.8cm} 
>{\centering\arraybackslash}p{0.8cm} 
>{\centering\arraybackslash}p{1.0cm} 
>{\centering\arraybackslash}p{0.8cm} 
>{\centering\arraybackslash}p{1.1cm} 
>{\centering\arraybackslash}p{0.8cm} 
>{\centering\arraybackslash}p{0.8cm} 
>{\centering\arraybackslash}p{0.8cm} 
>{\centering\arraybackslash}p{0.9cm}}
\toprule
\textbf{Survey} & \textbf{Link} & \textbf{Year} & \textbf{Venue} & \makecell{\textbf{Data} \\ \textbf{Rep.}} & \makecell{\textbf{MM} \\ \textbf{LLM}} & \makecell{\textbf{Diff.} \\ \textbf{Model}} & \makecell{\textbf{Unif.} \\ \textbf{FW }} & \makecell{\textbf{Bench-} \\ \textbf{mark}} & \makecell{\textbf{Eval.} \\ \textbf{Metr.}} & \makecell{\textbf{Sub-} \\ \textbf{tasks}} & \textbf{App.} & \makecell{\textbf{chall-} \\ \textbf{enges}} \\
\midrule
The Visual Analysis of Human Movement: A Survey & \cite{22b} & 1999 & CVIU & \cmark & \xmark & \xmark & \xmark & \cmark & \cmark & \cmark & \xmark & \xmark \\
Recent Developments in Human Motion Analysis & \cite{22c} & 2003 & Patt Recogn & \cmark & \xmark & \xmark & \xmark & \xmark & \xmark & \xmark & \xmark & \xmark \\
A Survey of Advances in Vision-Based Human Motion & \cite{22d} & 2006 & CVIU & \xmark & \xmark & \xmark & \xmark & \cmark & \cmark & \xmark & \xmark & \xmark \\
Content-Based Management of Human Motion Data & \cite{22e} & 2021 & IEEE Acc. & \cmark & \xmark & \xmark & \xmark & \cmark & \cmark & \xmark & \xmark & \cmark \\
3D Human Motion Prediction: A Survey & \cite{20} & 2022 & Neuro comp. & \cmark & \xmark & \xmark & \xmark & \cmark & \cmark & \xmark & \xmark & \xmark \\
Deep GenAI Models on 3D Representations: A Survey & \cite{22a} & 2023 & arXiv & \cmark & \xmark & \xmark & \xmark & \xmark & \xmark & \xmark & \cmark & \cmark \\
Human Motion Generation: A Survey & \cite{21} & 2023 & TAMI & \cmark & \xmark & \cmark & \xmark & \cmark & \cmark & \xmark & \xmark & \cmark \\
MultiModal Generative AI: LLM, Diffusion & \cite{22} & 2024 & arXiv & \xmark & \cmark & \cmark & \cmark & \xmark & \xmark & \xmark & \xmark & \xmark \\
Human Motion Video Generation: A Survey & \cite{22f} & 2024 & arXiv & \cmark & \cmark & \cmark & \xmark & \xmark & \cmark & \cmark & \xmark & \cmark \\
Human Image Generation: A Comp Survey & \cite{19} & 2024 & ACM CS & \xmark & \xmark & \cmark & \xmark & \cmark & \cmark & \xmark & \cmark & \cmark \\
A Survey of Cross-Modal Visual Content Generation & \cite{18} & 2024 & CSVT & \xmark & \cmark & \cmark & \xmark & \cmark & \cmark & \xmark & \xmark & \cmark \\
Appearance and Pose-Guided Human Generation & \cite{22g} & 2024 & ACM CS & \cmark & \xmark & \cmark & \xmark & \cmark & \cmark & \cmark & \xmark & \xmark \\
Synthetic Data in Human Analysis: A Survey & \cite{22k} & 2024 & TPAMI & \cmark & \xmark & \xmark & \xmark & \cmark & \cmark & \cmark & \xmark & \xmark \\
Establishing a Unified Evaluation Framework for HMG & \cite{22z} & 2024 & arXiv & \xmark & \xmark & \xmark & \xmark & \xmark & \cmark & \xmark & \xmark & \xmark \\
Autoregressive Models in Vision: A Survey & \cite{22m} & 2024 & arXiv & \xmark & \cmark & \xmark & \xmark & \xmark & \cmark & \xmark & \xmark & \cmark \\
\textbf{Ours} & \textbf{--} & \textbf{2025} &--  & \cmark & \cmark & \cmark & \cmark & \cmark & \cmark & \cmark & \cmark & \cmark \\ 
\bottomrule
\end{tabular}
\end{table}

\subsection{Text Data Representations}

Textual representations are pivotal in Text to Motion \textbf{(T2M}) generation tasks, especially when employing text-conditioned models. Text is typically represented as a sequence of words or tokens, which are converted into numerical vectors using word embeddings or more advanced models, such as transformers. Traditional embedding techniques, like Word2Vec \cite{49} and GloVe \cite{50}, produce fixed, non-contextualized word representations, where semantically similar words are placed closer together in a high-dimensional space. In contrast, recent transformer-based models, such as BERT \cite{51} and GPT \cite{12}, generate contextualized embeddings, meaning the representation of each word is influenced by its surrounding context within a sentence, enabling richer, more nuanced semantic understanding.

For instance, in models like BERT~\cite{61}, the dimensionality of the embeddings typically ranges from 768 to over 1000, which allows the model to capture intricate relationships between words in different contexts. In this way, a sentence like ``A person is walking slowly'' is tokenized into individual words (``A,'' ``person,'' ``is,'' ``walking,'' ``slowly''), with each token represented as a vector in the embedding space. These vectors, such as ``walking'' $\rightarrow$ $[v_1, v_2, \ldots, v_D]$, where $D$ represents the embedding dimension, are learned by the model and capture the word’s semantic meaning relative to its context. These vectors are then used as input to models that generate corresponding motion sequences, ensuring that the generated movement aligns with the described action. The recent emergence of LLMs has significantly contributed to the evolution of text-to-motion generation models, making them more varied and refined \cite{52} \cite{53}. These LLMs excel in encoding rich semantic and contextual information, allowing them to guide motion generation models with greater precision, ensuring the generated motion corresponds more closely to the text description.

\subsection{Ways of Motion Data Collection}
Human motion data can be collected through four main methods: marker-based motion capture, which uses reflective markers for precise tracking;\textbf{ markerless} motion capture, which employs computer vision techniques for more flexible, natural settings; \textbf{pseudo-labeling}, where experts label data by hand, often used for fine-tuning but is labor-intensive. Each method has its advantages and is selected based on the specific needs of the task.
\textbf{Marker-based }motion capture involves placing reflective markers or Inertial Measurement Units (IMUs) on specific points of a person's body to track movement in 3D space. This data can be processed using techniques such as forward kinematics or parametric body meshes (e.g., \textbf{SMPL}) for detailed motion representation \cite{33, 54, 55}. Optical markers offer high precision but are limited to indoor settings, while IMUs are portable and usable outdoors. \textbf{Markerless }motion capture, on the other hand, relies on computer vision algorithms and multiple synchronized cameras to track body movement without the need for markers, making it more flexible but less accurate \cite{56, 57, 58}. Pseudo-labelling is another technique used for monocular RGB images or videos, where 2D or 3D key points are estimated using models such as OpenPose or VideoPose3D; however, it tends to be less precise than motion capture systems \cite{59,60,61}. Lastly, manual annotation involves creating human motion by hand through animation tools, a labor-intensive process that delivers high-quality results but lacks scalability.

\section{ Foundational Multimodal LLM }

Multimodal Large Language Models (\textcolor{blue}{MLLMs}) have recently gained prominence and are now leading advancements in visual understanding, particularly in tasks related to motion generation. In this section, we will review and discuss the recent research in the cross-domain of \textcolor{blue}{GenAI} and \textcolor{blue}{MLLM}. Though the discussion of MLLM is paramount, we first consider the \textcolor{blue}{LLM} and its nature of work.

\subsection{Autoregressive Probabilistic LLM}

\textcolor{blue}{LLM} is the primary module in MLLM that receives input in various forms, including user queries in the form of text, visual information, and descriptions, and responds to the query after it has been interpreted. The nature of LLM is \textbf{autoregressive} - it predicts the next word based on the sequence of previous words, and it can be shown in the equation Eq. (1).

\begin{equation}
    P(z) = \prod_{t=1}^{T} P(z_t \mid z_1, z_2, \cdots, z_{t-1}).
     \tag{\textcolor{blue}{1}}
\end{equation}

In this formula, $z$ represents a sequence of tokens, where each token $z_t$ is predicted based on the previous tokens $z_1, z_2, \cdots, z_{t-1}$. This autoregressive approach, used in transformer-based \textcolor{blue}{LLMs}, enables step-by-step generation of sequences, where each token $z_t$ depends on its preceding tokens.

A primary consideration is how LLMs can handle visual information, such as human motion, since they typically process text tokens as their primary input. To integrate visual data, recent works \cite{53,61a,63, 69,70} align textual descriptions with both text and visual encoders from \textcolor{blue}{Vision Language Modeling (VLM)} during pre-training. A prominent example of VLM is CLIP \cite{61a,108}.

Some recent approaches represent motion sequences as a set of discrete visual tokens, treating visual motion as a ``foreign language'' \cite{64} through the use of codebooks and vector quantization. This setup enables the use of both text and visual tokens, which can be processed in an autoregressive manner by LLMs.

In the following subsections, we will discuss text tokenizers, \textcolor{blue}{VLMs}, and visual tokenizers, with a focus on recent work in motion understanding and generation.

\subsection{Text Tokenizer }

Text tokenization converts raw text into smaller units, such as subwords, words, or characters, which are then mapped to numerical representations for model processing. Traditional tokenization methods, such as \textbf{Byte Pair Encoding (BPE)} and WordPiece, efficiently handle text sequences but struggle with capturing fine-grained motion semantics \cite{13,51}. To bridge this gap, attention mechanisms and embedding strategies play a crucial role in text-based encoders by capturing contextual dependencies and ensuring semantic richness in multimodal tasks \cite{104}. Self-attention enhances long-range dependencies, while embedding layers transform tokenized inputs into dense representations, facilitating better alignment between textual descriptions and motion features \cite{67,92,147}. Recent advancements explore hierarchical and continuous tokenization approaches to further refine these representations, ensuring improved granularity and contextual awareness for motion reasoning and synthesis.

\subsection{Visual Tokenizer }

\begin{figure}
    \centering
    \includegraphics[width=1.0\linewidth]{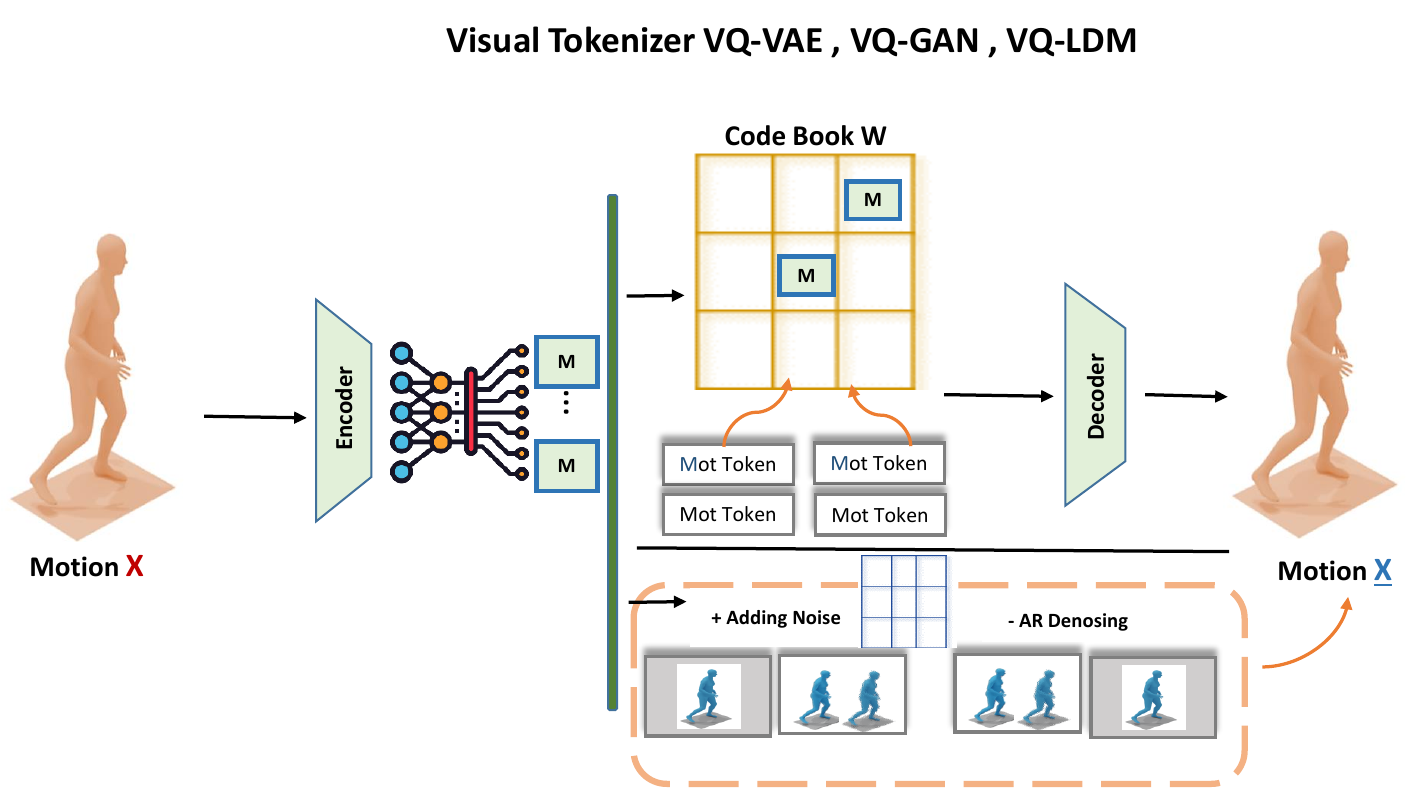}
    \caption{The architecture of a visual tokenizer typically used in generative AI. The model converts visual inputs (e.g., images or video frames) into discrete tokens that can be processed by language or multimodal models, enabling effective cross-modal generation.}
    \label{FIG:2}
\end{figure}

The mechanism of \textcolor{blue}{Visual Tokenizer (VT)} is to convert video frames or images into a discrete set of tokens that the model can process for generating motion similar to a text tokenizer in \textbf{NLP}. For the generation of motion, the \textbf{VT} breaks down motion sequences into basic units, which include body joint positions, limb movements, and key frames that enable the model to understand and generate human motion patterns \cite {13}. In early research on images or video frames, they are broken down into a series of patches and paired with a continuous linear embedding. The inspiration comes from LLM with language modeling. The discrete tokenizer gained fame for being able to transform motion frames into sequences of tokens \cite{61}. Fundamentally, visual tokenizers based on vector quantization with the codebook concept are VQ-VAEs \cite{14,17,67}, VQ-GAN \cite{92,95}, and VQ-Diffuse \cite{101,121,146}. We simplify the concept of motion generation using each model and illustrate it in Figure~\ref{FIG:2}. 
\subsubsection{VQ-VAE }

In the context of motion understanding and generation, \textbf{VQ-VAE} \cite{6,17} operates similarly to an auto-encoder, consisting of an encoder $F(\cdot)$ and a decoder $G(\cdot)$. For a given motion sequence $y$, VQ-VAE first transforms it into a lower-dimensional continuous representation $F(y)$ using the encoder. This encoded representation is then discretized via a codebook $W = \{w_m\}_{m=1}^M$, which serves as a set of discrete motion prototypes, akin to word embeddings in language models. Each prototype $w_m \in \mathbb{R}^{d_m}$ represents a specific motion feature. Recent research has focused on architectural enhancements to VAEs, leading to improved variants such as VQ-VAE \cite{82,92,93,94,95} and advancements in backbone structures with Residual Vector Quantization (RVQ-VAE) \cite{97,98,99,100}. These improvements contribute to generating more realistic motion while reducing computational complexity.

To convert the continuous vector into a discrete token, the closest match between the encoded vector $F(y)$ and the codebook entries is found using:
\begin{equation}
    \text{Discrete}(F(y)) = w_r, \quad r = \arg\min_r ||F(y) - w_r||
     \tag{\textcolor{blue}{2}}.
\end{equation}

Once this discrete token $w_r$ is determined, it is used to reconstruct the motion sequence via the decoder $G(w_r)$, generating the output $\hat{y} = G(w_r)$.

The training loss function for VQ-VAE in motion generation is as follows:
\begin{equation}
    \begin{aligned}
        L = ||y - G(w_r)||^2 &+ ||\text{stopgrad}[F(y)] - w_r||^2 
        + \alpha ||\text{stopgrad}[w_r] - F(y)||^2.
    \end{aligned}
    \tag{\textcolor{blue}{3}}
\end{equation}
\begin{itemize}
    \item The first term ensures that the reconstructed motion $G(w_r)$ closely matches the original motion sequence $y$.
    \item The second term pushes the motion token $w_r$ to align with the encoded vector $F(y)$.
    \item The third term pulls the encoded vector $F(y)$ towards the selected discrete token $w_r$.
\end{itemize}

This objective enables VQ-VAE to efficiently transform motion sequences into discrete tokens, thereby facilitating their effective use in generating and reconstructing motion sequences.

\subsubsection{VQ-GAN }

In the context of motion generation, \textbf{VQ-GAN} \cite{75} employs a GAN-based perceptual loss instead of the traditional $L_2$ reconstruction loss, enabling the model to learn a more expressive and detailed codebook. Recent advancements in GAN variants have demonstrated comparable performance in motion tokenization, styling, and reconstruction \cite{81,83}. Despite its drawbacks, such as mode collapse, GANs offer significantly faster training times compared to diffusion models and present substantial opportunities for architectural improvements. The tokenization process for motion generation can be illustrated through an example.

Given a motion sequence $z$ with dimensions $T \times M \times D$, where $T$ is the time steps, $M$ represents the number of joints or key points, and $D$ is the feature dimension, the encoder $P(\cdot)$ compresses it into a lower-dimensional vector representation $P(z)$ of size $t \times m \times d_c$, where $t < T$, $m < M$, and $d_c$ is the codebook dimension. This results in $t \times m$ vectors, each with dimensionality $d_c$.

For each encoded vector, the nearest neighbor in the codebook $Q = \{q_n\}_{n=1}^N$, where $q_n \in \mathbb{R}^{d_c}$, is identified for discretization:
\begin{equation}
    \text{Discrete}(P(z)) = q_k, \quad k = \arg\min_k ||P(z) - q_k||.
     \tag{\textcolor{blue}{4}}
\end{equation}

This process produces a discrete sequence of length $t \times m$, representing the motion in a compressed form. The decoder $G(q_k)$ then reconstructs the motion sequence using the selected discrete tokens:
\begin{equation}
    \hat{z} = G(q_k).
     \tag{\textcolor{blue}{5}}
\end{equation}

The GAN perceptual loss encourages high-quality reconstructions, ensuring that the generated motion sequence $\hat{z}$ not only resembles the original motion $z$ but also contains detailed and realistic movement dynamics, capturing complex motion patterns. This makes VQ-GAN well-suited for tasks like motion generation, where detailed temporal and spatial coherence is required.

\subsubsection{VQ-Diffuse}

\textbf{VQ-Diffuse} functions as a generative model that combines diffusion processes with visual tokenization techniques. The process begins with the encoder $E(\cdot)$, which takes an input image $x$ of size $H \times W \times 3$ and maps it into a lower-dimensional latent representation $E(x)$ of size $h \times w \times n_c$, where $h < H$ and $w < W$. This latent representation consists of $h \times w$ tokens, each with dimension $n_c$.

To discretize these latent vectors, \textbf{VQ-Diffuse} \cite{121} employs a codebook $Z = \{ z_k \}_{k=1}^K$, analogous to word embeddings in NLP. For each latent vector $E(x)$, the model identifies the nearest token in the codebook, resulting in a discrete representation $z_q$. This is achieved through the following process:
\begin{equation}
    z_q = \text{Discrete}(E(x)) = z_k, \quad k = \arg\min_k \| E(x) - z_k \|.
     \tag{\textcolor{blue}{6}}
\end{equation}

The discrete tokens are then subjected to a diffusion process \cite{121a,121b}, which iteratively refines the representation to generate high-quality outputs. The training objective of VQ-Diffuse can be framed as:
\begin{equation}
    L = \sum_{t=1}^{T} \left( \| x - D(z_q^{(t)}) \|^2 + \lambda \cdot \text{Loss}_{\text{diffusion}} \right),
     \tag{\textcolor{blue}{7}}
\end{equation}
where $D(\cdot)$ represents the decoder that reconstructs the image from the discrete tokens, $\lambda$ is a balancing hyperparameter, and $\text{Loss}_{\text{diffusion}}$ represents the diffusion loss that promotes smooth transitions between the generated frames.

By utilizing this approach, VQ-Diffuse effectively captures the complex dynamics of motion through a refined understanding of visual tokens. It enables the generation of motion sequences that are not only coherent and realistic but also contextually relevant to the visual content they represent, thereby advancing motion generation capabilities and reducing computational cost.
\begin{figure}[t]
    \centering
    \includegraphics[width=1.1\linewidth]{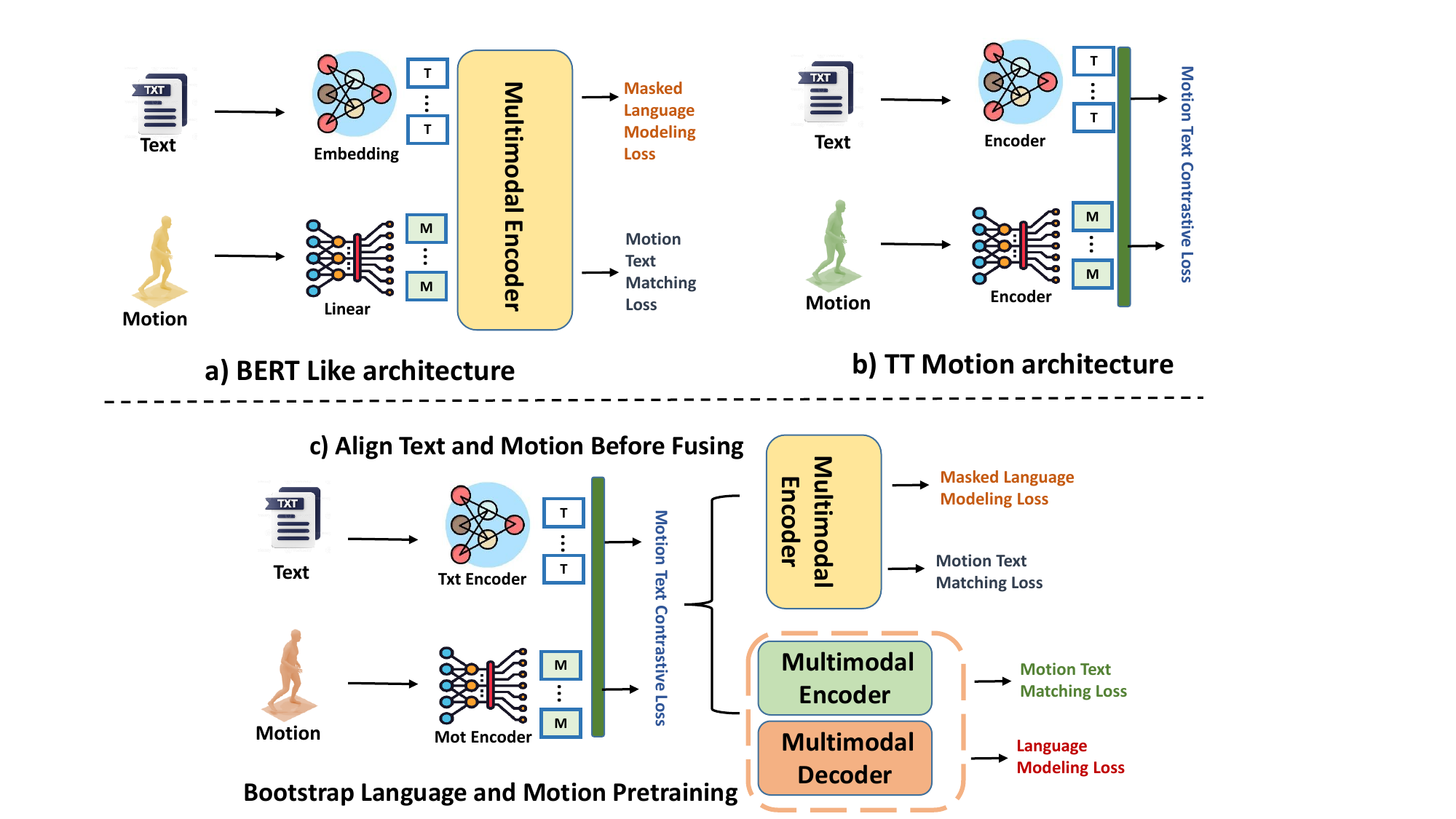}
    \caption{An illustration of various architectures used in motion understanding and generation, including \textbf{a)} BERT-like architectures, \textbf{b)} contrastive learning frameworks, and \textbf{c)} CLIP-based pretraining models.}
   \label{FIG:4}
   \label{FIG:4a}
   \label{FIG:4b}
   \label{FIG:4c}
\end{figure}

\subsubsection{Analysis of LLM with VQ-VAE , VQ-GAN and VQ-Diffuse}

In the context of motion generation, VQGAN and VQ-VAE act as effective visual tokenizers, converting motion data—such as videos or sequences of key points representing human movement—into discrete tokens. This transformation allows the data to be integrated into LLMs, facilitating a comprehensive understanding of both visual and textual information. By encoding motion sequences into a format compatible with LLMs \cite{61}, these models enable a range of applications, including generating coherent motion sequences from textual descriptions and enhancing the realism of animated characters in various environments.
Furthermore, VQGAN and VQ-VAE provide powerful mechanisms for compressing motion data into lower-dimensional representations, which can be utilized in Latent Diffusion Models (LDMs). By working in a latent space, these models \cite{7, 67,92, 95, 101, 121, 146} can introduce and refine motion characteristics through the addition and removal of noise, ultimately generating dynamic sequences from text prompts. The ability to represent complex motion patterns as discrete tokens significantly enhances the capacity of LLMs to model intricate relationships between language and movement, allowing for more nuanced motion generation and prediction tasks.
The integration of visual tokenization methods with LLMs creates a robust framework for addressing challenges in motion generation, enabling the generation of realistic motion sequences from both textual and visual inputs. This synergy not only streamlines the generation process but also opens avenues for applications such as virtual reality, animation, and human-computer interaction, where understanding and producing fluid motion is crucial.

\subsection{Vision Language Modeling (VLM)}

A \textcolor{blue}{\textbf{VLM} }is an \textcolor{blue}{AI} model that integrates large-scale visual and textual data to perform tasks such as understanding, retrieving, and generating content across both modalities. It combines visual perception, such as motion understanding, with Natural Language Processing (NLP) to generate descriptions of visual content \cite {13}. The success of NLP models like BERT \cite{51} has influenced pretraining and fine-tuning methods, driving advancements in multimodal \textcolor{blue}{AI}. Key components shown in Figure~\ref{FIG:4} of \textbf{VLMs} include vision encoders, language encoders, and multimodal fusion, with applications in motion captioning, Visual Question Answering (VQA), motion-text matching, motion-to-image generation, and cross-modal retrieval \cite{61a,89,107,108,144}. Popular models, including CLIP, ViLT, and Flamingo, have been utilized in the task of \textbf{HMUG}.


\subsection{Multi-Modal LLM in \textbf{HMUG}}

After discussing the architectures of the \textcolor{blue}{MLLM} –we will discuss how the current work in motion generation based on these architectures. The researchers have conducted comprehensive work on image generation using LLM. Afterward, they started working on human motion and video generation. Specifically, in human motion generation, it requires a sequence of image frames; some motion LLMs also utilize audio, speech, and scene modalities. Though our focus is on LLM text-to-motion generation. Motion LLM has higher computational complexity compared to other text modalities, such as image LLMs. Due to the considerable challenge of collecting high-quality motion datasets for training, early fusion techniques are often viewed as computationally intensive in the context of human motion and video generation. Most of the existing work relies on alignment methods, such as cross-attention and contrastive learning, which aid in synchronizing heterogeneous data representations, thereby enhancing coherence in multimodal generation tasks. The architecture of both techniques is illustrated in detail in Figure~\ref{FIG:4a}a.

\begin{figure}[t]
    \centering
    \includegraphics[width=1\linewidth]{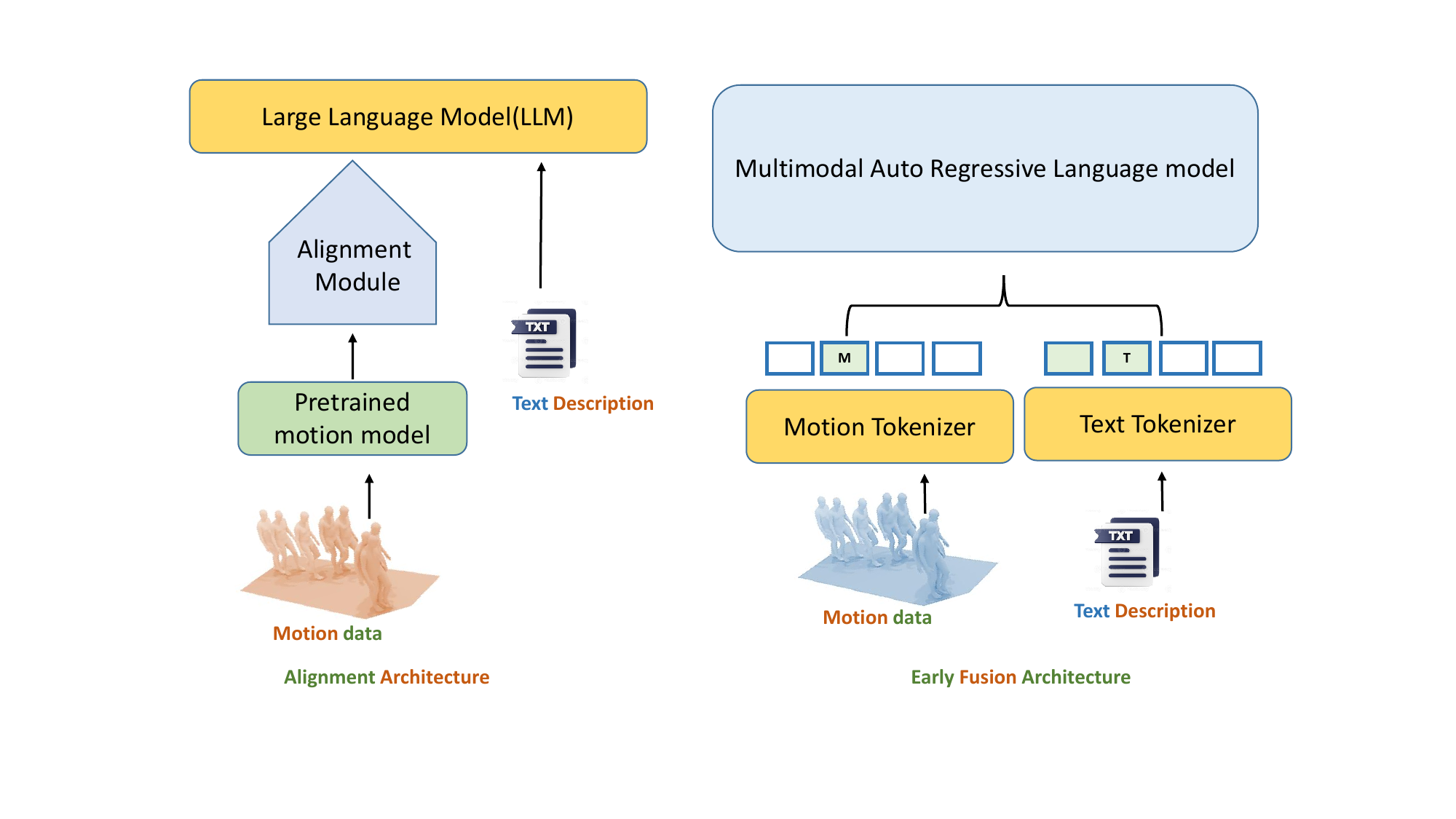}
    \caption{A representation of early fusion and alignment architectures employing contrastive learning. The early fusion approach combines multimodal features at the input or intermediate level, while the alignment architecture projects modalities into a shared embedding space, facilitating cross-modal contrastive learning.
}
    \label{FIG:5}
\end{figure}

Recently, researchers shared their work on Motion-Agent \cite{97}, A dialogue-based framework for generating human motion from textual descriptions. They used a simple adapter to fine-tune only 1-3 percent of the Motion LLM \cite{52} modal parameters with pre-trained GPT-4 and achieve comparative results. The base model, MotionLLM \cite{62}, is designed for understanding human behavior from both human motions and videos. They introduced the MoVid benchmark dataset, bridging the gap between motion and video modalities. Both MotionLLM and MotionAgent showed fantastic results in generation, understanding, captioning, and reasoning. The limitations faced by these researchers are on video encoders. Another work proposed is the AvatarGPT \cite{63} framework, impressed by the base work InstructGPT \cite{117}. This work constructed a dataset and achieved SOTA results in low-level tasks, showing promising results in high-level tasks. This work serves as the interface for task planning, decomposition, generation, summarization, and understanding. 

Another work proposed by the authors, MotionGPT \cite{64}, and declared improved diversity and realism by GPT-3 prompting. They also transform the high-level text to low-level text using zero-shot prompting techniques. Another work on fine-tuned LLM with Motion general-purpose motion generators, MotionGPT \cite{65}. They used a multi-control signal, which includes text and single-frame poses, instead of the single modality of the control signal. This is the uniqueness of their work. They also employed 
Low-Rank Adaptation \textbf{(LORA}), similar to motionAgent, by freezing most of the LLM parameters, and obtained comparable results. These techniques are efficient for generating motion based on conditions. Recently, authors proposed MotionScript \cite{66}, an LLM-based approach for generating expressive 3D human motion descriptions. They also introduced a novelty in motion text conversion and NLP representations of human motion. The primary contribution of this work is the automatic generation of detailed captions from motion using an LLM, achieving a higher level of granularity than T2M-GPT \cite{67}. 

On the other hand, a groundbreaking study, MotionChain \cite{71} and TAAT \cite{72}, leverages LLMs, vision-language models, and human motion data to enhance motion generation tasks. The authors introduced a multimodal conversational motion controller that utilizes multimodal prompts for virtual humans. The emergence of text-to-motion generation has opened up numerous compelling applications, attracting significant research interest. However, Alert-Motion \cite{69} raises security concerns regarding the integration of LLMs with motion generation, highlighting the risk of malicious exploitation through adversarial inputs injected into black-box text-to-motion models. This risk is further amplified when such malicious motion generation is applied to humanoid controllers. To address these challenges, researchers propose LLM-based agents that iteratively refine and search for adversarial prompts, employing contrastive modules to ensure semantic and contextual relevance.

Currently, researchers are exploring real-world applications of human motion generation, including robotics \cite{77,78}, autonomous vehicles \cite{79,79a}, and healthcare \cite{80}. A recent study presented at the Intelligent Vehicle Symposium, Walk the Talk \cite{73}, introduced an LLM-based pedestrian motion generation model. The authors constructed a dataset that captures pedestrian motion behavior while crossing roads, thereby contributing to advancements in autonomous driving simulations. Another notable study on humanoid motion generation \cite{76}, published by Intel Labs, explores its applicability in virtual agents. Additionally, healthcare applications are emerging, such as an LLM-driven fitness coach chatbot \cite{64}, demonstrating the potential of motion-aware \textcolor{blue}{AI} assistants. Table~\ref{tab:2} presents a comparative study of LLMs on downstream tasks related to understanding and generation.
\renewcommand{\arraystretch}{1.2}
\begin{table}[t]
\caption{A comprehensive comparisons of technical papers in multimodal LLM in motion understanding and generation, showing the modalities being used, representations, backbone, and datasets. It also highlights the tasks and subtasks catered to in the research, with open-source availability. T: Text; M: Motion ; AR: Autoregressive.}
    \centering
    \label{tab:2}
    \resizebox{\linewidth}{!}{%
    \footnotesize
    \begin{tabular}{>{\arraybackslash}p{1.8cm} >{\centering\arraybackslash}p{0.6cm} >{\centering\arraybackslash}p{1.6cm} >{\centering\arraybackslash}p{1.2cm} >{\centering\arraybackslash}p{1.5cm} >{\centering\arraybackslash}p{0.7cm} >{\centering\arraybackslash}p{1.3cm} >{\centering\arraybackslash}p{1.5cm} >{\centering\arraybackslash}p{1.9cm} >{\centering\arraybackslash}p{0.7cm} >{\centering\arraybackslash}p{1.0cm}}
    \toprule
        \textbf{Method} & \textbf{Link} & \textbf{Submission Date} & \textbf{Input Modality} & \textbf{Multimodal LLM} & {\textbf{Motion Rep.}} & \textbf{Backbone} & \textbf{Datasets} & \textbf{Downstream Tasks} & {\textbf{Open Source}} & \textbf{Venue} \\ 
    \midrule
        MotionLLM & \cite{52} & 27-May-2024 & T \& M & GPT-4, VQ-VAE, Vicuna-7B & 3D & Encoder Decoder & BABEL &  Generation, Captioning, Reasoning & \textcolor{green}{\checkmark} & ArXiv \\
        
        AvatarGPT & \cite{63} & 28-Nov-2023 & T & Lama-13B, GPT2-large, T5 & 3D  & Auto Regressive & HumanML3D &  Generation, Understanding, Summarization & \textcolor{green}{\checkmark} & CVPR \\
        
        MotionGPT-3 & \cite{65} & 19-Jun-2023 & T & GPT-3, DistilBERT & 2D, 3D & Encoder Denoiser & HumanML3D, BABEL & Retrieval, Generation & \textcolor{green}{\checkmark} & CVF \\
        
        MotionGPT-2 & \cite{116} & 29-Oct-2024 & T \& M & LLMs, VQ-VAE & 3D & Encoder Decoder & HumanML3D, Motion X, KIT-ML &  Generation, Captioning, Prediction & \textcolor{red}{\xmark} & ArXiv \\
        
        T2M-GPT & \cite{64} & 24-Sep-2023 & T \& M & VQVAE, GPT & 3D  & AR Encoder Decoder & HumanML3D, KIT-ML & Reconstruction, Generation & \textcolor{green}{\checkmark} &  CVF \\
        
        MotionChain & \cite{68} & 2-Apr-2024 & T \& M & ChatGPT, VQ-VAE & 3D & Encoder Decoder & TMR, HumanML3D & Generation, Reasoning, Translation & \textcolor{red}{\xmark} & ECCV \\
        
        MotionScript & \cite{66} & 19-Dec-2023 & M & GPT-4.0 & 3D & Encoder Decoder & HumanML3D, BABEL &  Description Generation & \textcolor{green}{\checkmark} & CVPR \\
        
        WalkLLM & \cite{194} & 5-Jun-2024 & T \& M & Flan-T5, VQVAE & 2D & Encoder Decoder & HumanML3D, KIT-ML & Pedestrian Generation & \textcolor{red}{\xmark} & IVS \\
        
        MotionLLMVid & \cite{62} & 30-May-2024 & T \& M & GPT-4, LLaVA & 3D & Vision Encoder Decoder & HumanML3D, Motion-X, H3DQA & Understanding, Reasoning, Generation & \textcolor{green}{\checkmark} & ArXiv \\
        
        PRO-Motion & \cite{60a} & 22-Dec-2023 & T & GPT-3.5, DDPM & 2D, 3D & Auto Regressive & HumanML3D, Motion-X, PoseScript & Planning, P-Diffuser, Generation & \textcolor{red}{\xmark} & ECVA \\
        
        FineMoGen & \cite{73} & 22-Dec-2023 & T & LLM, Diffusion & 3D & Encoder Decoder & MoGen, HumanML3D, KIT-ML & Generation, M-Editing & \textcolor{green}{\checkmark} & NeurIPS \\
        
        AlertMotion & \cite{69} & 1-Aug-2024 & T \& M & GPT-3.5, T2M & 3D & Encoder Decoder & HumanML3D, AMASS & Adversarial Similarity, Naturality & \textcolor{red}{\xmark} & ArXiv \\
    \bottomrule
    \end{tabular}}
\end{table}

\section{ Preliminaries in Multimodal Diffusion }
In this section, we will first provide an outline to key concepts in human motion generation before delving into diffusion models. Human motion generation techniques can generally be grouped into two categories. The first category comprises regression-based methods, which utilize supervised learning to directly predict motion from input features by mapping conditions to target motions. The second category is based on generative models, which aim to model the underlying motion distribution or its joint distribution with input conditions, typically using unsupervised learning. Prominent generative models include Generative Adversarial Networks (\textbf{GANs}), Variational Autoencoders (\textbf{VAEs}), Normalizing Flows (\textbf{NFs}), and Denoising Diffusion Probabilistic Models (\textbf{DDPMs}). Additionally, specialized models, such as motion graphs, are frequently used in areas like computer graphics and animation.

\begin{figure}[t]
    \centering
    \includegraphics[width=1\linewidth]{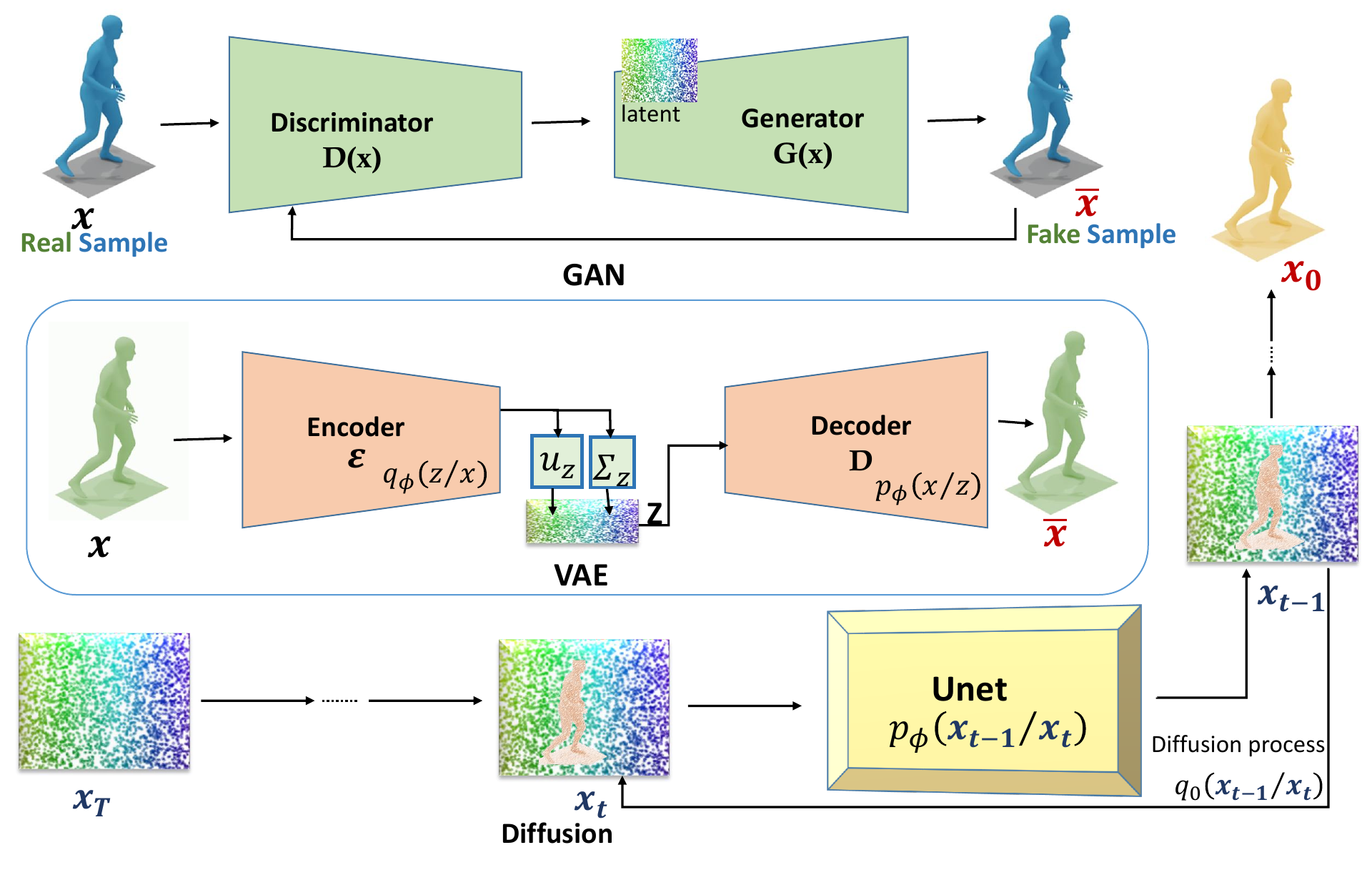}
    \caption{ An illustration of generative model architectures, including GAN, VAE, and Diffusion models. }
    \label{FIG:6}
\end{figure}

With this background, we will introduce the above generative models Figure~\ref{FIG:5}, diffusion probabilistic modeling, detailing its framework along with available variants of the models as shown in Figure~\ref{FIG:6}. We will also discuss the latent diffusion model and its significance, as well as advanced models for generating text-to-motion outputs.

\textbf{Generative Adversarial Networks.}
Generative Adversarial Networks (GANs) \cite{74} have gained popularity for human motion generation due to their capability to model complex data distributions. In this framework, GANs consist of two neural networks—a generator (\(G\)) and a discriminator (\(D\))—that engage in an adversarial process. The generator’s objective is to produce realistic human motion sequences from a random noise vector, \(z\), while the discriminator's role is to differentiate between real human motion data and the synthetic motion generated by the generator (\(G\)).The adversarial setup can be framed as a min-max game, with its objective captured by the following loss functions:
\begin{equation}
    \min_G \max_D \left( 
    \mathbb{E}_{x \sim p_{\text{data}}} \log D(x) 
    + \mathbb{E}_{z \sim p_z} \log(1 - D(G(z))) 
    \right).
        \tag{\textcolor{blue}{8}}
\end{equation}

\textbf{Discriminator's Loss}
\begin{equation}
    L_D = -\mathbb{E}_{x \sim p_{\text{data}}} \log D(x) 
          - \mathbb{E}_{z \sim p_z} \log(1 - D(G(z))) 
          \tag{\textcolor{blue}{9}},
\end{equation}

\textbf{Generator's Loss}
\begin{equation}
    L_G = -\mathbb{E}_{z \sim p_z(z)} \left[ \log D(G(z)) \right].
        \tag{\textcolor{blue}{10}}
\end{equation}

In these equations, $p_{\text{data}}(x)$ represents the real data distribution (i.e., the distribution of human motion), and $p_z(z)$ represents the noise distribution (typically a Gaussian distribution). The generator aims to minimize $L_G$, meaning it seeks to create realistic motion sequences that maximize the discriminator’s belief that they are real. Meanwhile, the discriminator attempts to maximize $L_D$, learning to more effectively differentiate between real and generated motion sequences. The architecture of the GAN model is represented in Figure~\ref{FIG:6}.

GANs \cite{89a} have been successfully applied to generate various types of human motion \cite{73,75}, such as walking, running, and complex dance sequences. Some advanced GAN architectures incorporate temporal features using techniques like Recurrent Neural Networks (RNNs) or 3D convolutional networks, enabling them to handle the time-dependent nature of motion data. For example, models like Temporal GAN (TGAN) \cite{76, 77} extend traditional GANs by incorporating 3D spatial-temporal convolutions to generate continuous motion over time, thereby improving realism. Additionally, variants like Deep Convolutional GAN (DCGAN) \cite{78} enhance the modeling of spatial features in motion, Progressive Growing GAN (PGGAN) \cite{79} allows for refined output through gradual complexity, and StyleGAN \cite{80,81} provides control over different aspects of generated motion, enabling the creation of diverse and stylized sequences.

\textbf{Remark.} Despite their success, GANs in motion generation face challenges such as mode collapse, where the generator produces limited varieties of motion, and training instability due to the delicate balance between  $G$ and $D$. Recent work has focused on addressing these challenges. Still, they remain key areas of ongoing research.

\textbf{Variational Autoencoders.} VAEs have emerged as a prominent approach for generating human motion sequences, offering a robust framework for representing and reconstructing complex motion data. Unlike GANs, which rely on adversarial training, \textbf{VAEs} \cite{88,89,90} utilize an encoder-decoder architecture to learn a probabilistic latent space, ensuring that the latent variables \( z \) are drawn from a smooth and continuous distribution, typically a Gaussian, shown in Figure~\ref{FIG:6}. 

In motion generation, the encoder transforms human motion data \( x \) into a latent representation \( z = E(x) \), and the decoder reconstructs the motion sequence \( \hat{x} = D(z) \), aiming to closely approximate the original motion data.

VAEs employ a unique training objective that incorporates the Evidence Lower Bound (ELBO) to balance reconstruction accuracy and regularization of the latent space. The loss function can be expressed as:
\begin{equation}
    L(\alpha, \beta; x) = - D_{KL}(q_\beta(z|x) \| p_\alpha(z)) 
    + \mathbb{E}_{q_\beta(z|x)} \log p_\alpha(x|z).
    \tag{\textcolor{blue}{11}}
\end{equation}

This optimization ensures that the generated motions preserve high fidelity, while also enforcing the latent space to adhere to a known distribution, thereby reducing the risk of overfitting.

\textbf{Remark.} A common challenge faced by VAEs is the phenomenon of posterior collapse, where the latent variables become less informative, relying too heavily on the decoder. Despite this limitation, VAEs and their variants, such as Conditional VAE (CVAE) \cite{9} and Vector Quantized VAE (VQ-VAE) \cite{85,94}, have proven effective in capturing both static and dynamic aspects of human motion, making them valuable tools for generating diverse and realistic motion sequences.
\begin{figure*}[t]
    \centering
    \includegraphics[width=1\linewidth]{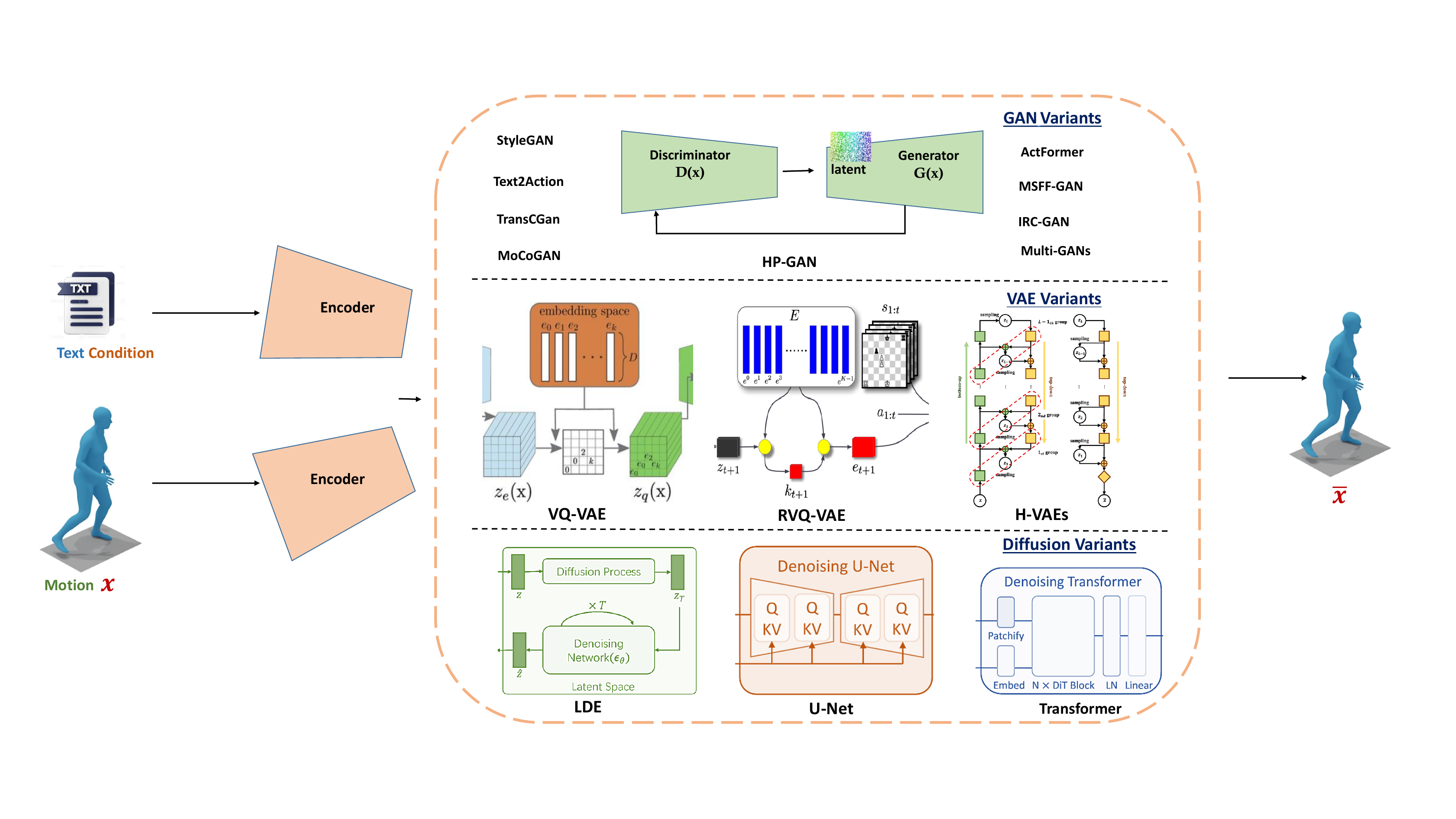}
    \caption{ An illustration of multimodal diffusion GAN and VAE variants in the architecture. The figure highlights how each variant is integrated into the architecture, with the diffusion GAN enabling generative learning across multiple modalities and the VAE variants facilitating effective latent space modeling for multimodal data synthesis and transformation.}
    \label{FIG:7}
\end{figure*}
\\
\textbf{Diffusion Probabilistic Modeling.} Diffusion models have shown great promise in generating human motion \cite{4a,62}, leveraging their probabilistic framework to produce realistic motion sequences through a gradual denoising process, as illustrated in Figure~\ref{FIG:5}. The core idea revolves around a forward diffusion process, where Gaussian noise is added incrementally to human motion data over a series of iterations. This noise is regulated by a predefined schedule and leads to a set of noisy motion sequences. The goal is to learn how to reverse this process, beginning with random noise and gradually denoising it until realistic human motion is generated. The forward diffusion process can be formalized as:
\begin{equation}
    q(x_t \,|\, x_{t-1}) = \mathcal{N}(x_t; (1 - \beta_t)x_{t-1}, \beta_t I),
      \tag{\textcolor{blue}{12}}
\end{equation}
where \( x_t \) represents the motion data at step \( t \), and \( \beta_t \) controls the amount of noise added at each step. As the number of steps \( T \) approaches infinity, the motion data becomes indistinguishable from gaussian noise. However, the key to generating realistic motion lies in the reverse process. Given the noisy motion data, a neural network \( p_\theta \) learns to predict the previous less noisy step, gradually recovering the clean motion:
\begin{equation}
    p_\theta(x_{t-1} \,|\, x_t) = \mathcal{N}(x_{t-1}; \mu_\theta(x_t, t), \Sigma_\theta(x_t, t)).
        \tag{\textcolor{blue}{13}}
\end{equation}

This process generates human motion by iteratively refining noisy sequences, and the model is trained using an objective similar to that of Variational Autoencoders (VAEs) \cite{82}, which minimizes the evidence lower bound (ELBO).

When applied to human motion, diffusion models can generate smooth and continuous sequences like walking or running by training on large datasets of motion data. They provide high-quality results and stable training compared to other generative models, such as GANs, though they can be computationally expensive due to the extended sequence of reverse steps required.

To tackle the computational cost, Latent Diffusion Models (\textbf{LDMs}) \cite{9} offer a more efficient approach by conducting the diffusion process in a compressed latent space, rather than directly on motion sequences. This reduces the dimensionality of the data and speeds up the generation process while maintaining motion realism. This latent space can be compressed using an encoder, similar to VQGAN \cite{83}, allowing for conditional generation where additional inputs, such as text or semantic information, can guide the motion generation.

In essence, diffusion models are a powerful tool for generating high-fidelity human motion, though optimizing their efficiency continues to be a key focus of research.
\renewcommand{\arraystretch}{1.2}
\begin{table}[H]
\caption{A comprehensive comparisons of technical work in multimodal LLM in motion understanding and generation. Showing the modalities being used, representations, backbone, and datasets. It also highlights the tasks and subtasks catered to in the research, with open-source availability.}
\centering
\label{tab:3}
\resizebox{\linewidth}{!}{%
\footnotesize
\begin{tabular}{p{1.5cm} p{0.9cm} p{0.7cm} p{1.8cm} p{1.5cm} p{1.5cm} p{1cm} p{1.5cm} p{2.5cm} p{1cm} p{1cm}}
\toprule
\textbf{Method} & \textbf{Link} & \textbf{Year} & \textbf{Input} & \textbf{GenAI Model} & \textbf{Represen-tation} & \textbf{Backbone} & \textbf{Datasets} & \textbf{Tasks} & \textbf{Open source} & \textbf{Venue} \\
\midrule
Text2Motion & \cite{87} & 2018 & Text + Motion & GAN & 2D, 3D & Enc-Dec & MSR-VTT & Text-2-Motion, Motion-2-Text & \xmark & ICRA \\ 
HP-GAN & \cite{88} & 2018 & Text + Motion & WGAN & 3D & Enc-Dec & NTU, Human3.6M & Motion Prediction & \xmark & CVPR Wksp. \\ 
GAN model & \cite{88}   & 2021 & Text + Motion & GAN, LSTM, BERT & 2D & Enc-Dec & KIT-ML & Quality Motion Gen. & \cmark & ICCV \\ 
ActFormer & \cite{78} & 2023 & Text & GAN, Transformer & 2D, 3D & Enc-Denoiser & GTA, NTU, BABEL & Multi-Person Gen. & \cmark & ICCV \\ 
Style-GAN & \cite{86} & 2024 & Text + Motion & GAN & 3D & Autoreg. Enc-Dec & HumanML3D, KIT-ML & Motion Generation & \xmark & ICCGV \\ 
LS-GAN & \cite{89} & 2024 & Text + Motion & WGAN-GP & 3D & Enc-Dec & HumanAct12, HumanML3D & Motion Synthesis & \xmark & arXiv \\ 
MSFF-GAN & \cite{90} & 2024 & Motion & MSFF-GAN & Keypts, 3D & Enc-Dec & Human Video & Motion Transfer & \xmark & CCDC \\ 
\cdashline{1-11}
MotVAE  & \cite{85} & 2023 & Text + Motion & VQ-VAE, Trans. & 3D & Enc-Dec & KIT-ML, HumanML3D & Motion Generation & \cmark & CVPR \\
VAE DLS & \cite{94} & 2024 & Text + Motion & VQ-VAEs & Keypts, 2D & Enc-Dec & KIT-ML & Motion Reconstruction & \xmark & ICMEW \\ 
\cdashline{1-11}
Tevet & \cite{62} & 2023 & Text + Motion & Diffusion, Trans. & 2D, 3D & Enc-Dec & HumanML3D, KIT-ML & Interp., Editing, Recog. & \cmark & ECCV \\ 
FineMoGen & \cite{76}  & 2023 & Text+ Motion & LLM, Diffusion & 3D & Enc-Dec & HuMMan, KIT-ML & Motion Gen., Editing & \cmark & NeurIPS \\ 
ReMoDiffuse & \cite{112} & 2023 & Text+ Motion & Retrieval Diffusion & 3D & Enc-Dec & HumanML3D, KIT-ML & Hybrid Retrieval & \cmark & ICCV \\ 
MotionDiffuse & \cite{4a} & 2024 & Text + Motion & Diffusion, Trans. & 2D, 3D & Text Enc, Diffuse & HumanML3D, KIT-ML, BABEL & Text-to-Motion & \cmark & TPAMI \\ 
MotionFix & \cite{113} & 2024 & Text+ Motion & Diffusion & 3D & Trans.-Denoiser & HumanML3D, KIT-ML, PoseFix & Motion Gen., Editing & \cmark & SIG- GRAPH \\ 
\bottomrule
\end{tabular}
}
\end{table}

\subsection{Text Conditioned Motion Generation }
Text is a powerful and intuitive medium, enabling the specification of human actions, making it a compelling modality for motion sequence synthesis compared to other inputs, such as predefined motion categories or voice commands. Text-to-motion generation leverages the descriptive nature of language, capturing complex aspects of human movement, including action types, velocities, directions, and destinations. This flexibility enables rich, nuanced control over motion synthesis, allowing for diverse applications in animation, robotics, and healthcare. The task can be divided into two main categories: action-to-motion, where the text directly maps to predefined actions, and text-to-motion, which aims to dynamically generate realistic motion sequences from free-form language descriptions, as shown in Table~\ref{tab:3}.

 \subsubsection{Word-based Action to Motion} 
The action-to-motion task \cite{90a,90b,91} focuses on generating human motion sequences based on predefined action categories, such as ``Walk'' and ``Jump''. These actions are typically represented using simple encoding methods, such as one-hot vectors, making it easier to translate specific actions into motion. This method \cite{125,130,142} offers a direct and efficient approach to motion generation, providing more control and predictability compared to text-to-motion tasks, which deal with the complexities of interpreting natural language. However, action-to-motion generation is more rigid, often limited to individual actions or short sequences. It struggles to handle more complex motions involving multiple actions, detailed trajectories, or nuanced body movements. To overcome these limitations, researchers have introduced the concept of using localized action signals to guide the generation of more complex, global motion sequences, allowing for smoother transitions and more accurate representations of human movement \cite{72}, actformer \cite{75}.

\subsubsection{Text-description to Motion} 
The text-to-motion task involves generating human motion sequences directly from natural language descriptions, tapping into the rich expressiveness of language \cite{61}. Unlike action-to-motion, which relies on a fixed set of predefined labels, text-to-motion allows for the creation of a wide range of dynamic and complex motions based on diverse textual inputs. This approach enables more flexible and nuanced motion generation but also presents challenges in translating detailed linguistic descriptions into accurate physical movements, requiring a deep understanding of both language and human motion dynamics. We will discuss each text-to-motion generative model, along with its available variants, one by one, as shown in Figure~\ref{FIG:6}.
\subsection{T2M Using GAN with G-variants}

Now, after discussing the architecture of GAN in section 4.1 and illustrating it in Figure~\ref{FIG:5}. We explore text-to-motion generation using GANs with G-variants, which involves generating human motion sequences from text by leveraging \textbf{GANs} with various generator architectures. These \textbf{G-variants} enable the system to produce more realistic and diverse motions by improving the quality of generated sequences. This approach addresses challenges such as motion realism and fine-tuning complex actions based on textual descriptions, as illustrated in Figure~\ref{FIG:6}. 

The researchers proposed Text2Motion \cite{84}, in which they utilized a GAN architecture to generate diverse motion from high-level natural language descriptions. For example, GAN body pose utilizes a CNN for detection and generates high-quality human body poses through a GAN. Another study generates a motion dataset from a textual description using a GAN model \cite{85}. One of the research areas is decomposing motion \cite{5} to generate video content. The authors proposed ActFormer \cite{75}, which employs a GAN and transformer to generate action-conditioned motion based on text. It helps by adapting the diverse motion representations in single-person motion generation tasks. They also introduced synthetic data for multi-person behavior. In terms of prediction, HP-GAN \cite{16} proposed a probabilistic 3D human motion work by modifying the Wasserstein GAN (WGAN) model. They introduced the novel loss function and predicted the next motions non-deterministically. Style-GAN \cite{83} generates an improved human motion generation by interpolating the intermediate latent variables. In TransCGan \cite{86}, the researcher used the transformer architecture with a conditional GAN to generate human motions. MSFF-GAN \cite{87} proposed the multiscale feature fusion GAN for motion transfer by using global-local perceptual loss and pose consistency loss. 

\subsection{T2M Using VAE with V-variants} 

After discussing the architecture of VAE in Section 4.2 and illustrating it in Figure~\ref{FIG:5}, we can see that text-to-motion generation is achieved using Variational Autoencoders (\textbf{VAEs}) \cite{88,89}. It is the most commonly used \textcolor{blue}{GenAI} model that transforms natural language descriptions into human motion sequences by learning a compressed latent representation of motion data. VAEs \cite{90} capture the complex variability in human motion and generate outputs by sampling from this learned space, conditioned on text. This enables the model to produce realistic, diverse motions with smooth transitions while reducing noise and enhancing overall quality \cite{91}.
Advanced VAE variants like VQ-VAE \cite{82,92,93,94,95} and most recent RVQ-VAE \cite{17,96,97,98,99,100,101,102,103} further enhance the process by introducing discrete latent spaces. \textbf{VQ-VAE} produces sharper, more coherent outputs. For instance, AttT2M \cite{104} proposed a body part and global-local two-stage attention mechanism utilizing a discrete latent space. In contrast,\textbf{ RVQ-VAE} employs multiple levels of quantization to achieve more detailed and fine-tuned motions. Other variants, such as Conditional VAEs (\textbf{CVAEs}) \cite{9} and Hierarchical VAEs (\textbf{HVAEs}) \cite{27,107}, enhance control over motion generation by incorporating features like text conditioning and multi-level latent representations, enabling more intricate and precise motion generation.  HumanTOMATO \cite{105} uses the hierarchical VAE to generate a vivid and well-aligned whole-body motion generation through textual descriptions.

\subsection{T2M Using Diffusion with D-variants}

In recent years, diffusion models have demonstrated remarkable success in text-to-image generation, prompting researchers to extend these models to the domains of text-to-motion and video synthesis. By adapting attention mechanisms, researchers have successfully achieved text-to-motion generation without requiring the introduction of additional parameters. The pioneering work L2Pose \cite{106} addresses multimodal integration challenges by constructing a joint embedding space characterized by shorter and more tractable sequences. MotionDiffuse \cite{4a} represents a seminal contribution in this area, drawing significant attention for enabling high-quality text-to-motion generation using a probabilistic diffusion framework. This model employs a cross-modality linear transformer that incrementally integrates text into the motion sequence, thereby mitigating the dominance of text features within the joint embedding space and its associated high-dimensional latent representations. Given the strong correlation between natural language commands and human body movements, this approach effectively controls individual body parts.

A related effort added by FLAME \cite{109}, with a focus on high fidelity motion generation that aligns well with the given text prompt. The combination of these architectures ultimately yields an efficient and robust model for generating diverse and high-quality motion. Another researcher \cite {110} uses a classifier-free diffusion model with BERT embedding and achieves comparable results in complex NLP scenarios. He showcased the result in zero-shot motion generation through unseen text guidance. The guided and controllable motion synthesis \cite{111} using the diffusion model provides a new direction for incorporating spatial constraints into the generation process. They injected the sparse key frame signals that are susceptible to being missed in the reverse denoising step. Another area of interest is representation learning in latent space, a technique that offers efficient modeling. The Motion Latent Diffusion (\textbf{MLD}) \cite{9} generates reasonable human motion sequences that are compatible with text descriptions or action classes. While comparing this MLD model to Motion Diffusion Model (\textbf{MDM}), there is very little need for computational overhead to generate vivid motions, and it is also faster than MDM and more effective, as shown in the architecture in Figure~\ref{FIG:6}. These controllable motion generation methods are generally classified into two types: classifier-based and classifier-free guidance, as outlined in previous studies.

MotionDiffuse is grounded in the Denoising Diffusion Probabilistic Model (\textbf{DDPM}) \cite{107}, which excels at producing diverse outputs while accommodating additional constraints throughout the denoising process. Building on this foundation, Tevet \cite{108} introduces a transformer-based diffusion model that adapts classifier-free guidance for the human motion domain. Similarly, FLAME \cite{109} focuses on high-fidelity motion generation aligned with textual prompts, contributing to the development of robust models capable of synthesizing diverse and high-quality human motions.

Additional work includes a classifier-free diffusion approach that integrates BERT embeddings, achieving competitive performance in complex natural language processing tasks, including zero-shot motion generation from unseen text prompts \cite{110}. Furthermore, research in guided and controllable motion synthesis has proposed mechanisms for injecting spatial constraints during generation, such as sparse keyframe signals, which are particularly vulnerable to omission during the reverse denoising phase \cite{111}.

An emerging area of interest involves representation learning in the latent space for more efficient motion modeling. The MLD model \cite{9} is capable of generating coherent motion sequences corresponding to text descriptions or action classes, with lower computational overhead and faster inference compared to MDM models, as illustrated in Figure~\ref{FIG:6}. These controllable generation methods are generally categorized into classifier-based and classifier-free guidance strategies, as outlined in earlier studies.

Some approaches concentrate on generating motion from fine-grained textual descriptions. For instance, FineMoGen \cite{73} introduces a spatiotemporal fine-grained motion generation method supported by the HuMMan-MoGen dataset, which also facilitates motion editing—a promising advancement in human motion manipulation. The issue of ensuring consistency between text and motion is addressed by ReMoDiffuse \cite{112}, which combines retrieval-based techniques with diffusion and transformer architectures and introduces a novel generalizability metric. More recently, MotionFix \cite{113} introduced a text-annotated dataset for unconstrained human motion editing. By training a conditional diffusion model on this dataset, the approach outperforms existing state-of-the-art methods. Likewise, GUESS \cite{114} proposes a cascaded latent denoising diffusion model that incrementally enriches motion generation by adaptively inferring joint conditions. Through semantic graph modeling, hierarchical information is retrieved to support fine-grained generation, with semantic disentanglement performed across motions, actions, and specific details.

Broadly, existing research employs a two-pronged intervention strategy to enhance motion generation: one at the textual level to improve descriptive quality and another at the motion level through joint space optimization. The overarching objective across these efforts is to establish a generation pipeline that is efficient, controllable, high-quality, and diverse, while minimizing both training and inference time.

\section{Unified Human Motion Model}

In the preceding sections, we explored the capabilities of multimodal LLMs and multi-model diffusion models in addressing various aspects of human motion planning, understanding, and generation. In this section, we present a unified framework that integrates these approaches, aligning with the broader trend toward Artificial General Intelligence (\textbf{AGI}). This unified approach envisions a single model that can plan, understand, and generate human motion seamlessly. We examine this framework from perspectives including model architecture and probabilistic methodologies.

\subsection{Multimodal Probabilistic Modelling}

Autoregressive models, as demonstrated in multimodal LLMs, excel in text understanding and generation due to their autoregressive nature. Similarly, diffusion models have proven indispensable for generating high-quality motion, as outlined in the prior sections. Recent advancements in research propose combining these two paradigms, leveraging the strengths of both to create a unified framework for understanding and generating human motion. For instance, by aligning motion data and text through a shared latent space, these models enable the synthesis of realistic motion sequences guided by textual descriptions.

\subsection{Autoregressive Models in Motion Generation}

Although diffusion models dominate in generating high-quality visual motion, autoregressive models also hold significant promise for text-conditioned motion generation. MotionLLaMA \cite{115}, MotionGPT-2 \cite{116}, MotionLLM \cite{52}, and InstructMotion \cite{117} employ autoregressive approaches by discretizing motion data into tokens through Vector Quantized GANs or VAEs. These tokens, combined with text tokens, are processed by multimodal LLMs to generate sequences. For decoding, the tokens are mapped back into motion data using VQGAN or VAE, enabling the simultaneous understanding and generation of human motion.

Despite significant advancements, several challenges remain. Autoregressive models rely heavily on visual tokenizers that compress pixel-level motion information, often at the cost of losing fine-grained details critical for understanding and generating realistic motion. Additionally, the causal attention mechanism inherent to autoregressive models complicates token sequencing, frequently necessitating large datasets to achieve generalization and diversity, as shown in Figure~\ref{FIG:9a}a. BAMM \cite{115} addresses these issues to some extent by introducing a novel framework that can capture rich bidirectional dependencies and learn the probabilistic mapping between text inputs and motion outputs.

\subsection{Unify AR and Diffusion Model Capability}

\begin{figure}[t]
	\centering
	\includegraphics[width=1\columnwidth]{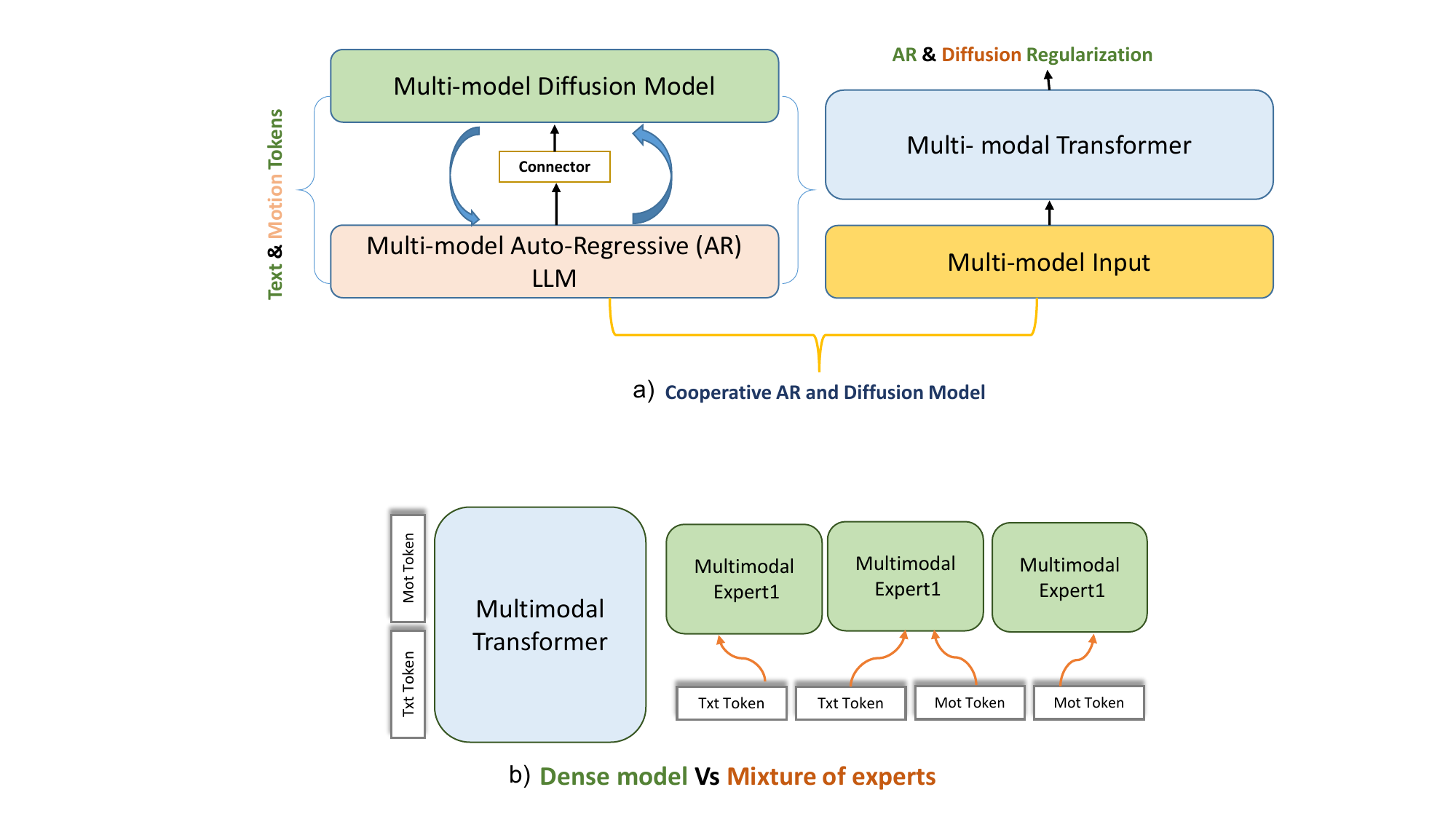}
	\caption{An illustration of \textbf{a)} the cooperative capabilities of an AR and Diffusion Model in a single unified framework, and \textbf{b)} the Dense versus Mixture of Experts alignment methodologies.}
	\label{FIG:8}
    \label{FIG:8a}
    \label{FIG:8b}
\end{figure}

To address these limitations, a unified framework combining the strengths of diffusion and autoregressive models is proposed, as shown in Figure~\ref{FIG:8a}a. Diffusion models, renowned for generating high-quality, plausible motion, can be integrated with autoregressive models to enhance understanding and generation. Existing approaches often use multimodal LLMs as controllers, with diffusion models handling motion generation. However, these methods may falter when text conditions fail to represent explicit motion details.

A more advanced unification strategy \cite{196} involves training both diffusion models and multimodal LLMs within a shared, high-dimensional embedding space shown in Figure~\ref{FIG:9a}a. This alignment enables the generation of realistic motion and understanding, conditioned on multimodal embeddings. Despite its potential, challenges such as independent modeling and alignment inconsistencies remain. A promising solution is the adoption of a transformer-based multimodal framework, such as AMD \cite{118}, complemented by an improved model AAMD \cite{119}, which processes both modalities jointly using full attention mechanisms to enhance comprehension and generation capabilities, as shown in the architecture in Figure~\ref{FIG:9c}c. MotionVerse \cite{120} introduced a Large motion Model (LMM) framework with ArtAttention, which incorporates a multimodal transformer and diffusion backbone, as shown in Figure~\ref{FIG:9d}d. This approach demonstrated superior knowledge exploitation using diverse training data, exhibiting strong generalizability. Similarly, M2D2M \cite{121} and DART \cite{122} have demonstrated promising results in handling long textual descriptions and enhancing motion controllability with improved realism.

\subsection{Unified Model Architecture}

The unified model architecture is designed to address multiple tasks, including comprehension, planning, understanding, reasoning, captioning, and motion generation. This architecture diverges from single-task approaches by incorporating joint pixel-level and semantic information, overcoming limitations in current theoretical concepts and visual motion generation techniques.

\subsubsection{Multimodal Input Architecture}

 As shown in Figure~\ref{FIG:8}, text is tokenized using standard methods, while motion data requires enhanced processing. Instead of relying solely on visual encoders for pixel-level inputs, a hybrid approach is proposed that combines pixel-level and semantic encoders to generate dual token representations. This flexibility enables adaptive token selection, thereby optimizing fine-grained understanding and generation tasks. Incorporating a mixture-of-experts approach further refines this process, as discussed in related literature.
\begin{figure*}[t]
    \centering
    \includegraphics[width=1.0\linewidth]{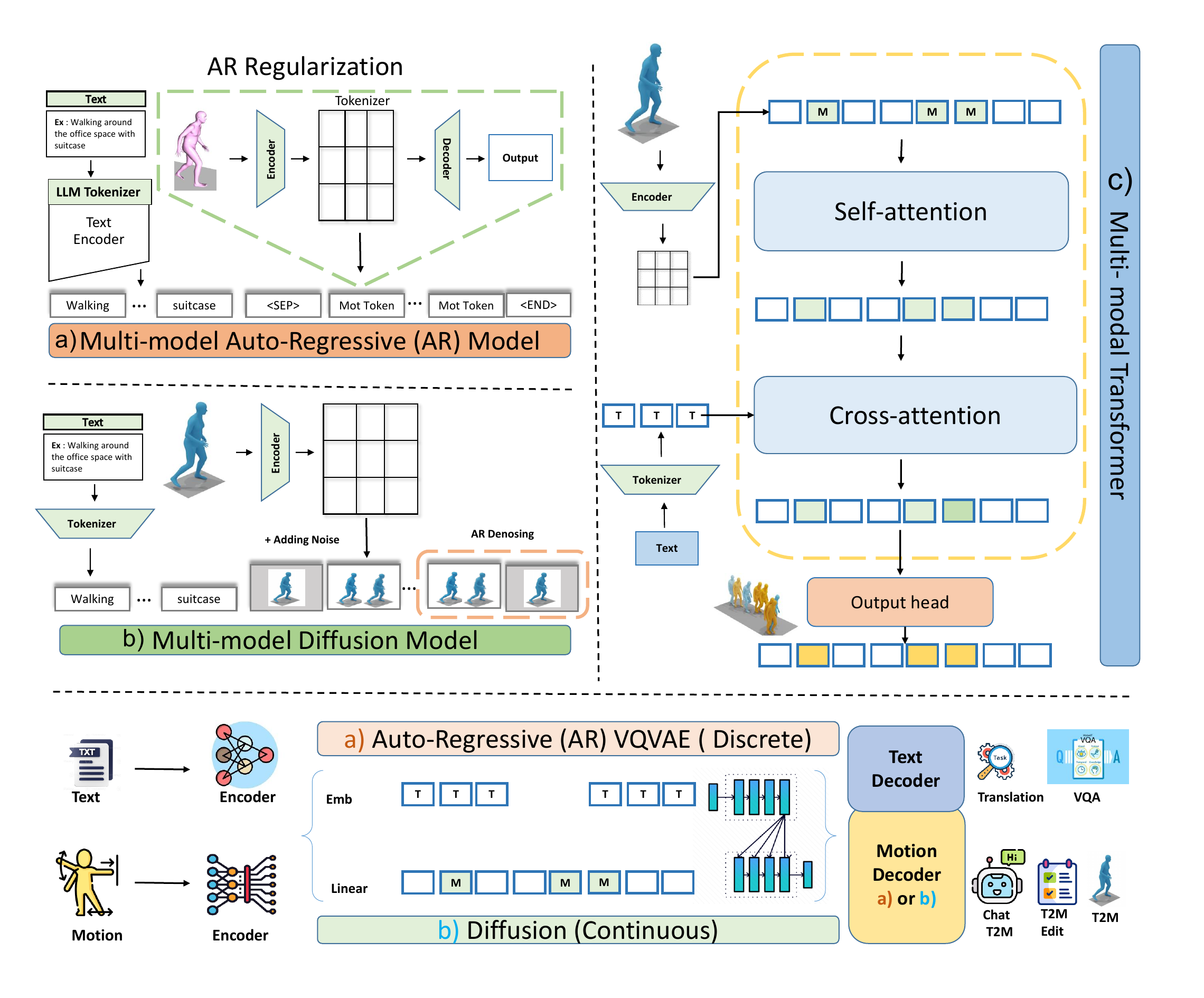}
    \caption{A detailed illustration of \textbf{a)} multimodal LLM, \textbf{b)} multimodal Diffusion, \textbf{c)} multimodal Transformer, and unified approaches with an encoder-decoder backbone architecture.}
    \label{FIG:9}
      \label{FIG:9a}
        \label{FIG:9b}
          \label{FIG:9c}
            \label{FIG:9d}
            
\end{figure*}

\subsubsection{Multimodal Transformer Architecture }

 The multimodal transformer captures complex relationships between text and motion modalities \cite{123,100,117}. Two configurations are illustrated in Figure~\ref{FIG:8b}b. The first involves a dense model serving as a unified transformer for both modalities, promoting shared knowledge. The second employs a mixture-of-experts strategy, where specialized components handle reasoning for understanding and generation tasks. This approach balances the strengths of autoregressive and diffusion models, ensuring robust performance across modalities. The unified architecture can be visualized in Figure~\ref{FIG:9}.

This section outlines a unified framework for multimodal \textbf{HMUG}, focusing on probabilistic modeling, input processing, and a transformer-based backbone architecture that addresses both multimodal diffusion and LLMs. By addressing the limitations of current methods and integrating advanced techniques, we aim to inspire further research into innovative multimodal architectures that bridge the gap between understanding and generation in human motion. Table~\ref{tab:4} presents a detailed unified architecture and fusion mechanism for both understanding and generation.

\renewcommand{\arraystretch}{1.2} 
\begin{table}[H]
\caption{Summary of related works on Unified text to human motion understanding, generation, and other subtasks. This table presents key details of recent studies, including their input modalities, methods, backbone models, motion representations, datasets, tasks and applications, open-source availability, and publication dates.}
\label{tab:4}
\centering
\resizebox{\linewidth}{!}{%
\footnotesize
\begin{tabular}{p{2.7cm} p{0.7cm} p{1.3cm} p{0.8cm} p{1.0cm}  p{1.4cm} p{1.5cm} p{1.5cm} p{1.5cm} p{0.7cm} p{1.1cm}}
    \toprule
    \textbf{Paper Title} & \textbf{Link} & \textbf{Method} & \textbf{Venue} & \textbf{Input Modality} & \textbf{Backbone Model} & \textbf{Motion Representation} & \textbf{Dataset} & \textbf{Tasks Applications} & \textbf{Open Source} & \textbf{Online Date} \\
    \midrule
    A Unified 3D Human Motion Synthesis Model via Conditional VAE & \cite{130a} & Unified 3D, HumanMotion & ICCV & Text, Motion & Conditional Variational Auto-Encoder & 2D, 3D  & Human 3.6M, CMU Mocap & Human pose series generation & \xmark & 28 Feb 2022 \\
    Pretrained Diffusion Models for Unified Human Motion Synthesis & \cite{123} & Mo-Fusion & CVPR & Text, Motion, Music & Transformer, DDPM, DistilBERT & 3D Rot. & KIT, AMASS, AIST++, LaFAN1, BABEL & Interactive motion editing and generation & \cmark & 06 Dec 2022 \\
    UDE: A Unified Driving Engine for HMG & \cite{148} & UDE Engine & CVPR & Text, Audio, Motion & VQ-VAE, Diffusion Decoder & 3D Rot. & Human ML3D, AIST++ &  Motion quantization and generation & \xmark & 24 June 2023 \\
    A Unified Framework for Multimodal, Multi-Part Human Motion Synthesis & \cite{130b} & Unified Motion & ArXiv & Text, Motion, Speech, Music & Hierarchical VQ-VAE, HuBERT, Transformer, ED & 3D Rot. & Human ML3D, AIST++, BEAT, MANO & Multi-part human motion generation & \cmark & 28 Nov 2023 \\
    MotionLLM: Multimodal Motion-Language Learning with LLMs & \cite{52} & Motion LLM & ArXiv & Text, Audio, Motion & RVQ, LLaMA, LoRA, Gemma, AR Learning & 3D Rot. & AIST++, Human ML3D & Single and multi-human motion generation & \cmark & 27 May 2024 \\
    AAMDM: AR Motion Diffusion Model & \cite{119} & AAMDM & CVPR & Text, Motion & AR,  Motion Diffusion Model & 3D Rot. & LaFAN1, AAMDM & Human motion generation and polishing & \xmark & 21 Aug 2024 \\
    LMM for Unified Multimodal Motion Generation & \cite{120} & LMM, MotionVerse, ArtAttention & ECCV & Text, Audio, Motion & Transformer, Diffusion Model & 3D Rot. & Human ML3D, AMASS, 3DPW & Motion Generation, Prediction & \cmark & 26 Oct 2024 \\
    MotionGPT-2: A General Purpose Motion Language Model for Motion Generation and Understanding & \cite{116} & Unified Motion GPT & ArXiv & Text, Motion & VQ-VAE, LLM, LoRA, T5, ED & 3D Rot. & Human ML3D, KIT-ML, MotionX & Human body-controlled motion generation & \xmark & 29 Oct 2024 \\
    MotionLLaMA: A Unified Framework for Motion Synthesis and Comprehension & \cite{115} & Motion LLaMA & ArXiv & Text, Audio, Motion & RVQ, LLaMA, LoRA, AR Learning & 3D Rot. & AIST++, Human ML3D, InterGen & Human motion synthesis and comprehension & \cmark & 26 Nov 2024 \\
    M2D2M: Multi-Motion Generation from Text with Discrete Diffusion Models & \cite{121} & DDM, VQ-VAE & ECCV & Text, Motion & Transformer, Diffusion, ED & 3D Rot. & Human ML3D, KIT-ML & Single and multimodal human motion generation & \xmark & 05 Dec 2024 \\
    \bottomrule
\end{tabular}}
\end{table}
\section{Datasets and Evaluation Metrics}

Following an in-depth discussion of multimodal \textbf{HMUG} architectures, along with their use cases, it is crucial to emphasize the role of datasets and evaluation metrics in implementing \textcolor{blue}{GenAI} models. This section explores commonly used text-to-motion datasets and evaluation metrics, structured into two subsections: Datasets and Evaluation Metrics.

Furthermore, many multimodal large models utilize these foundational datasets to construct task-specific datasets for downstream applications, such as motion captioning, conversational motion modeling, reasoning, and VQA, thereby enhancing their generalization capabilities. Table~\ref{tab:6} provides a summary of the key properties of these datasets, along with prominent state-of-the-art evaluation criteria, including newly developed datasets specifically designed for text-to-motion generation.

\section{Datasets and Evaluation Metrics}

Following an in-depth discussion of multimodal \textbf{HMUG} architectures, along with their use cases, it is crucial to emphasize the role of datasets and evaluation metrics in implementing \textcolor{blue}{GenAI} models. This section explores commonly used text-to-motion datasets and evaluation metrics, structured into two subsections: Datasets and Evaluation Metrics. Furthermore, many multimodal large models utilize these foundational datasets to construct task-specific datasets for downstream applications, such as motion captioning, conversational motion modeling, reasoning, and VQA, thereby enhancing their generalization capabilities. Table~\ref{tab:6} provides a summary of the key properties of these datasets, along with prominent state-of-the-art evaluation criteria, including newly developed datasets specifically designed for text-to-motion generation.

\subsection{Text and Motion Pairs Data}

\textbf{KIT-ML.} KIT Motion Language (KIT-ML) \cite{124} is a well-structured paired dataset designed for the integration of motion and text modalities. The motion data are captured using advanced optical marker-based systems, ensuring high precision in motion representation. Each motion sequence is paired with detailed language annotations that describe the associated movement. These annotations provide rich textual descriptions, making the dataset highly suitable for tasks such as motion understanding, generation, and text-to-motion mapping. KITML plays a crucial role in bridging the gap between natural language processing and motion analysis, enabling researchers to effectively develop and evaluate multimodal \textcolor{blue}{GenAI} models.

\textbf{BABEL.} The Bodies, Action, and Behavior with English Labels (BABEL) dataset \cite{130} extends the comprehensive AMASS \cite{131} motion dataset by providing rich text annotations for motion sequences. These annotations are available at two granular levels: sequence-level labels, which describe entire motion sequences, and frame-level labels, which offer insights into specific motion frames. BABEL encompasses over 28,000 motion sequences and 63,000 individual frames, spanning 250 diverse motion categories. This dual-level labeling makes BABEL an invaluable resource for tasks such as motion segmentation, classification, and text-to-motion generation, enabling more nuanced understanding and generation of human motion in \textcolor{blue}{GenAI} models.

\textbf{HumanML3D.} Human Motion Language 3D (HumanML3D) \cite{132}, a widely utilized dataset, is created by combining the HumanAct12 \cite{128} and AMASS \cite{131} datasets. It enriches motion sequences by associating each with three distinct text descriptions, enabling a deeper connection between textual inputs and 3D human motion. The dataset spans a diverse array of activities, including daily routines, sports, acrobatics, and artistic expressions. This diversity makes HumanML3D a valuable resource for advancing research in text-to-motion generation, offering robust benchmarks for aligning multimodal representations of text and motion.

\subsection{Motion Only Data}
\textbf{UESTC.} UESTC \cite{148} is a comprehensive dataset that captures motion data across three modalities: RGB videos, depth images, and skeleton sequences, all collected using the Microsoft Kinect V2 sensor. It is organized into 40 distinct action categories, with 15 categories that can be performed in both standing and sitting positions, and 25 categories exclusive to standing actions. The multimodal nature of UESTC makes it a valuable resource for tasks involving motion analysis, understanding, and generation, as it provides diverse representations of human actions in varying positions and contexts. Its detailed categorization supports the development and evaluation of advanced multimodal models for human motion.

\textbf{NTU-RGB+D 120.} The NTU-RGB+D 120 dataset  \cite{126} is an extended version of the widely-used NTU-RGB+D dataset \cite{127}, created to support more comprehensive research in \textbf{HMUG}. This extension introduces 60 additional action classes, bringing the total to 120 distinct categories that span a diverse range of activities, including daily routines and health-related tasks. Furthermore, it adds 57,600 new RGB+D video samples, significantly increasing the dataset's size and variety. Captured using multiple camera views and featuring subjects from different demographics, NTU-RGB+D 120 serves as a benchmark dataset for tasks like action recognition, motion generation, and multimodal reasoning, providing a robust foundation for generalization and real-world application.

\textbf{HumanAct12.} The HumanAct12 dataset \cite{128}, derived from the PHSPD\cite{129}, offers a curated collection of 3D motion clips, specifically designed to capture a diverse range of human actions. This dataset is structured around 12 main motion classes, such as walking, running, sitting, and warming up, which are further divided into 34 detailed sub-classes to represent nuanced variations of these behaviors. Each motion clip is carefully segmented, providing high-quality and detailed motion data suitable for tasks like motion understanding, generation, and action recognition. HumanAct12 is particularly valuable for applications that focus on everyday human activities, serving as a benchmark for exploring the capabilities of GenAI and multimodal models in understanding and replicating realistic human motion.

\subsection{Unified Data} 
Several foundational datasets have also been extensively used in combination for text-to-motion research, each contributing unique strengths to advance the field. \textbf{Human3.6M} \cite{133} is a large-scale motion capture dataset that features diverse human activities and accompanying annotations, making it instrumental in understanding motion dynamics. \textbf{CMU MoCap} \cite{134} provides a rich repository of skeletal motion data, encompassing a diverse range of activities, and offers a robust foundation for motion modeling. \textbf{AMASS} \cite{131} unifies over 15 motion capture datasets into a common parametric representation, enabling the creation of high-quality 3D motion sequences. \textbf{HuMMan} \cite{135}, on the other hand, is a more recent dataset designed to bridge the gap between \textbf{HMUG} by providing multimodal data including RGB, depth, and 3D motion.  Together, these datasets form a powerful foundation for building large-scale, diverse, and multimodal text-to-motion benchmarks, which support downstream tasks such as motion captioning, reasoning, and generative modeling. 

To achieve generalization and enhance the diversity of generated motions, several works have utilized the fusion of multiple datasets. \textbf{HUMANISE \cite{136}}, for instance, integrates scene, object, and textual data, with a total of 643 scenes. The dataset offers a rich combination of visual contexts and text annotations, enabling models to understand and generate motion sequences informed by complex interactions between objects, scenes, and associated text. In addition, \textbf{TED-Gesture \cite{137}} contains paired motion and text data, focusing on gesture-based motion generation. The inclusion of gestures and corresponding text descriptions enables models to learn the mapping between textual prompts and specific movements, supporting tasks such as text-to-gesture motion generation. Building upon this, \textbf{TED-Gesture++ \cite{138}} and \textbf{PATS \cite{139}} extend the fusion by introducing speech along with text and motion. These datasets enable the training of models to understand the impact of speech and text on motion, thereby enhancing the realism of generated gestures and motions in applications such as conversational \textcolor{blue}{AI} and gesture recognition. \textbf{BEAT \cite{140}} further complicates the relationship by incorporating emotional context along with text and speech, enabling models to capture how emotions can influence motion, such as body language or facial expressions. This multimodal approach, using datasets with diverse combinations of motion, text, speech, and emotion, allows large multimodal models to generalize better and produce more sophisticated and contextually aware motion sequences, leading to better performance in text-to-motion generation tasks.

\subsection{EVALUATION METRICS}

After discussing the datasets, it is essential to highlight the role of evaluation metrics, which are crucial for benchmarking the performance and quality of generative models in \textbf{HMUG}. These metrics serve to standardize comparisons across methodologies, enabling the field to progress. Assessing synthesized human motion is uniquely challenging due to its subjective nature, the intricacies of high-level conditional signals, and the one-to-many mapping between input and output. Broadly, evaluation focuses on aspects such as \textbf{fidelity}, \textbf{realism}, \textbf{diversity}, \textbf{coherence}, and \textbf{retrieval} performance. Metrics such as Fréchet Inception Distance (\textbf{FID}) are widely used to measure the distributional similarity between generated and real datasets. Table~\ref{tab:5} presents the various evaluation metrics used in human motion research, along with their interpretation, hyperparameters, bounds, and categories.

For assessing structural accuracy and motion quality, metrics such as Mean Per Joint Position Error (\textbf{MPJPE}) are employed. Retrieval metrics, such as R-Precision, are used to evaluate the model’s ability to accurately retrieve data. Perceptual evaluations, such as text-to-motion accuracy, provide insights into the alignment between generated motion and conditional input signals.

Diversity metrics, including Coverage, Recall, and Multimodality Score (\textbf{MS}), assess the variety and novelty of generated samples, ensuring the model captures diverse modes of human motion. Despite the diverse array of metrics available, establishing a unified evaluation framework for human motion synthesis remains a critical challenge \cite{22z}. For clarity, the evaluation metrics can be categorized based on their primary purposes in assessing generative models, such as fidelity, diversity, accuracy, and motion quality.
\subsubsection{Fidelity Metrics}
These metrics assess how closely the generated motion matches the real-world data, focusing on realism and naturalness:

\textbf{Fréchet Inception Distance (FID).} FID is a widely used metric for evaluating the quality of generative models. It measures the similarity between the real and generated data distributions by calculating the Fréchet Distance between their feature representations in a latent space \cite{10,94,64,114}. These features are typically extracted using a pre-trained neural network.

\begin{table}[t]
\caption{Summary of datasets and evaluation metrics used in various top-tier research with the backbone of Autoregressive LLMs, diffusion, and unified models for multimodal human motion understanding, planning, and generation, published in top-tier venues.}
\label{tab:5}
\centering
\renewcommand{\arraystretch}{1.2}
\setlength{\tabcolsep}{4pt}
\footnotesize
\begin{tabular}{p{1.2cm} p{1.3cm} p{1.8cm} p{1.8cm} p{4.5cm} p{4cm} }
    \toprule
\textbf{Metric} & \textbf{Category} & \textbf{Hyper-Parameters} & \textbf{Bounds} & \textbf{Interpretation} & \textbf{Used in Research} \\ 
\midrule
FID & Fidelity & None & $0 \leq \text{FID} < \infty$ & Lower is better; compares feature distributions with real data & \cite{10,64,94,114,153} \\ 
\hline
Precision & Fidelity &  T  Nearest points & $0 \leq \text{Precision} \leq 1$ & Closer to 1 indicates better fidelity &\cite{89,92,96,98,101,104,105}\\ 
\hline
Coverage & Diversity & T Nearest points & $0 \leq \text{Coverage} \leq 1$ & Higher indicate better diversity & \cite{4a,52,53,64,84,85,94,96,98,110,114}
 \\ 
\hline
Recall & Diversity  & T Nearest points & $0 \leq \text{Recall} \leq 1$ & Higher indicate better diversity & \cite{13,105,107,112}
 \\ 
\hline
MM Score & Diversity & T2M Accuracy & $0 \leq \text{MMS} \leq 1$ & Higher indicate better multi-modality representation & \cite{89,92,96,98,101,104,105,142,143,144,145,146,147}
 \\ 
\hline
MM Distrib. & Diversity & T2m Consistency & Depends on stat measures & Measures distribution similarity across modalities & \cite{13,101,122,140,141,142,143,144,145,146,147,148,149,150,151,152}
 \\ 
\hline
MPJPE & Accuracy & None & $0 \leq \text{MPJPE} < \infty$ & Lower indicate better accuracy in joint position estimation & \cite{88,99,118,109,95,119} \\ 
\hline
\end{tabular}
\end{table}

Proposed by \cite{153}, \textbf{FID} improves upon the Inception Score \cite{10} by directly comparing the distributions of real and generated samples. It assumes that both distributions follow a multivariate Gaussian distribution and calculates their means and covariances. The FID score quantifies the effort or ``energy'' required to transform one distribution into the other, where a lower FID indicates higher fidelity in the generated samples. Mathematically, FID is computed using:
\[
\text{FID}(P_1, P_2)^2 = \text{trace}(\Sigma_1 + \Sigma_2 - 2(\Sigma_1 \Sigma_2)^{1/2}) + \sum_{i=1}^{f} (\mu_{1,i} - \mu_{2,i})^2, \tag{\textcolor{blue}{14}}
\]
where \( \mu_1, \mu_2 \) are the means and \( \Sigma_1, \Sigma_2 \) are the covariance matrices of the feature representations of real and generated samples, respectively.

FID captures both the fidelity (how realistic the generated samples are) and diversity (how well they cover the real data distribution). However, it has limitations: a perfect FID score of 0 can be achieved if the model merely replicates the real data without producing novel samples. This makes it important to interpret FID in conjunction with other metrics or qualitative evaluations.

 \textbf{Precision (R-Precision).} It evaluates how well the generated samples conform to the real data distribution, focusing on fidelity. In the context of \textbf{HMUG}, it measures the proportion of generated motions that are close to the real motions in the feature space \cite{105, 104, 101, 98, 96, 92, 89}. Higher precision indicates that the AI-generated motions are realistic and align well with the characteristics of the real data. 

Mathematically, precision is often computed using a neighborhood-based approach, where a set of real samples and generated samples are compared in the latent space. For a given generated sample, if it falls within the neighborhood of any real sample (e.g., using k-nearest neighbors), it is considered precise. The metric is formally expressed as:
\[
\text{Precision} = \frac{\text{Gen samples in real neighborhood}}{\text{Total gen samples}}. \tag{\textcolor{blue}{15}}
\]
In motion generation, precision helps identify whether the model produces high-quality motions that are plausible when compared to real human motion. However, it does not assess diversity, meaning a high precision score could still occur if the model generates highly realistic but repetitive motions.

\subsubsection{ Diversity Metrics} These metrics assess the variety and novelty of the generated samples, ensuring that models do not suffer from mode collapse. In the context of \textbf{HMUG}, diversity metrics are crucial for evaluating whether the generated motions capture the full range of natural variations present in real human motion \cite{4a,94,53,52,64}.

\textbf{Coverage.} It measures how well the generated motions represent the diversity of the real motion data. It evaluates whether all major patterns or variations in real motion are reflected in the generated samples \cite{84,85,96,98,110,114}. Mathematically, coverage is often computed using a neighborhood-based approach in the latent space, where the proportion of real samples that are covered by the generated samples is calculated:
\[
\text{Coverage} = \frac{\text{Real samples in generated neighborhood}}{\text{Total real samples}}. \tag{\textcolor{blue}{16}}
\]
Higher coverage implies that the generative model captures a broader variety of real motion, ensuring that all significant motion patterns are adequately represented.

\textbf{Recall.} This metric quantifies the percentage of real motion samples that fall within the generated data distribution. In the context of motion generation, it measures the model's ability to recreate the diverse features of real human motion. Recall is calculated as:
\begin{equation}
\text{Recall} = \frac{\text{Real samples near generated}}{\text{Total real samples}}. \tag{\textcolor{blue}{17}}
\end{equation}

Closeness is typically measured using a distance threshold in a latent feature space (e.g., L2 distance or cosine similarity). A high recall score reflects the model’s ability to reproduce the diversity observed in real motion data. This metric is also used to evaluate a model’s understanding and retrieval of human motion~\cite{13,112}.

\textbf{Multimodality Score (MMS).} The MMS evaluates the extent of distinct modes or patterns present in the generated motions \cite{89,92,96,98,101,104,105,142,143,144,145,146,147}. It ensures that the model generates a variety of motion types, styles, or trajectories rather than repetitive or overly similar ones. Mathematically, MMS is computed as:
\[
\text{MMS} = \frac{1}{K} \sum_{k=1}^K \frac{\text{Unique Modes in Generated Motion} \, (G_k)}{\text{Modes in Real Motion Data} \, (R_k)}, \tag{\textcolor{blue}{18}}
\]
where \(K\) represents the number of motion categories or clusters, such as action types like walking, running, or dancing. For each motion category \(k\), \(G_k\) denotes the unique modes in the generated motion, referring to the distinct motion patterns produced. Similarly, \(R_k\) refers to the modes in real motion data, capturing the number of distinct motion patterns observed in the real-world dataset for the \(k^\text{th}\) category.
\renewcommand{\arraystretch}{2.9}
\begin{table}[H]
\caption{Comprehensive quantitative comparisons based on two important datasets: HumanML3D and KIT-ML. The table also describes the backbone architecture, tasks and subtasks. It highlights results including FID, Precision (Prec), Diversity (Div), and MM metrics.}
\label{tab:6}
\centering

\resizebox{\linewidth}{!}{%
\small
\begin{tabular}{p{2.1cm} p{2.9cm} p{1.5cm} p{2.5cm} p{4cm} p{1cm} p{1.5cm} | p{1cm} p{1cm} p{1cm} p{1cm} | p{1cm} p{1cm} p{1cm} p{1cm}}
\toprule
\textbf{Types} & \textbf{Method} & \textbf{Represen-tation} & \textbf{Backbone} & \textbf{Tasks} & \textbf{Open-source} & \textbf{Venue} & \multicolumn{4}{c|}{\textbf{HumanML3D \cite{132}}} & \multicolumn{4}{c}{\textbf{KIT-ML \cite{124}}} \\
& & & & & & & \textbf{FID} & \textbf{Prec} & \textbf{Div} & \textbf{MM} & \textbf{FID} & \textbf{Prec} & \textbf{Div} & \textbf{MM} \\
\midrule

\multirow{10}{*}{\textbf{\parbox{1.5cm}{Multimodal\\LLM}}}
& MotionLLM \cite{52}  & 3D &Enc-Dec & Motion Generation, Captioning, Multi-turn Editing, Reasoning, Composition & \textcolor{green}{\checkmark} & ArXiv & 0.491 & 0.521 & 9.838 & 3.138 & 0.781 & 0.433 & 11.407 & 2.982 \\
& AvatarGPT \cite{63} & 3D Latent & Autoregressive & Motion Generation, Understanding, Summarization & \textcolor{green}{\checkmark} & CVPR & 0.168 & 0.510 & 9.624 & - & 0.376 & - & - & - \\
& MotionGPT-3 \cite {65} & 2D, 3D & Enc-Denoiser & Retrieval, Generation & \textcolor{green}{\checkmark} & CVPR & 0.567 & 0.411 & 9.006 & 3.775 & 0.597 & 0.376 & 10.540 & 3.394 \\
& MotionGPT-2 \cite{116} & 3D & Enc-Dec & Motion Generation, Captioning, Prediction & \textcolor{red}{\xmark} & ArXiv & 0.232 & 0.492 & 9.528 & 3.096 & 0.510 & 0.366 & 10.350 & 2.328 \\
& T2M-GPT \cite{64} & 3D Latent & Autoregressive Enc-Dec & Reconstruction, Generation & \textcolor{green}{\checkmark} & CVPR & 0.116 & 0.491 & 9.761 & 1.856 & 0.514 & 0.416 & 10.920 & 1.570 \\
& MotionChain \cite{68} & 3D & Enc-Dec & Generation, Reasoning, Translation & \textcolor{red}{\xmark} & ECCV & 0.248 & 0.504 & 9.470 & 1.715 & - & - & - & - \\
& WalkLLM \cite{196} & Keypts, 2D & Enc-Dec & Pedestrian Motion Generation & \textcolor{red}{\xmark} & IVS & 0.197 & - & 8.157 & - & - & - & - & - \\
& FineMoGen \cite{73} & 3D & Enc-Dec & Generation, Editing & \textcolor{green}{\checkmark} & NeurIPS & 0.151 & 0.504 & 9.263 & 2.696 & 0.178 & 0.432 & 10.850 & 1.877 \\
& AlertMotion \cite{69} & 3D & Enc-Dec & Adversarial Similarity, Naturality & \textcolor{red}{\xmark} & ArXiv & 4.170 & 5.400 & 5.680 & - & - & - & - & - \\
& TAAT \cite{72} & 3D & Enc-Dec & Adversarial Similarity, Naturality & \textcolor{red}{\xmark} & ArXiv & 0.461 & 0.329 & 10.038 & 2.929 & 0.488 & 0.225 & 8.552 & 2.957 \\
\midrule

\multirow{8}{*}{\textbf{\parbox{1.5cm}{Multimodal \\ Diffusion}}}
& ActFormer \cite{75} & 2D, 3D & Enc-Denoiser & Single/Multi-person Gen. & \textcolor{blue}{\checkmark} & ICCV & 0.130 & 2.550 & - & - & - & - & - & - \\
& LS-GAN \cite{89} & 3D & Enc-Dec & Motion Synthesis & \textcolor{red}{\xmark} & ArXiv & 0.482 & 0.391 & 9.249 & 3.501 & - & - & - & - \\
\cdashline{2-15}
& VQ-VAE Mot \cite{85} & Keypts, 2D & Enc-Dec & Motion Recon. & \textcolor{red}{\xmark} & ICMEW & 0.017 & 0.505 & 9.532 & 3.005 & 0.251 & 0.417 & 10.750 & 3.143 \\
& VQ-VAE DLS \cite{94} & 3D & Enc-Dec & Motion Gen. & \textcolor{blue}{\checkmark} & CVPR & 0.141 & 0.492 & 9.722 & 1.831 & 0.514 & 0.416 & 10.921 & 1.570 \\
\cdashline{2-15}
& MotionDiffuse \cite{4a} & 2D, 3D & Enc-Denoiser & Text-to-Motion & \textcolor{blue}{\checkmark} & TPAMI & 0.630 & 0.491 & 9.410 & 1.553 & 0.514 & 0.416 & 10.921 & 1.570 \\
& Tevet \cite{108} & 2D, 3D & Enc-Dec & Interp., Editing, Recog. & \textcolor{blue}{\checkmark} & ECCV & 4.569 & 0.645 & 7.688 & 1.264 & 0.497 & 0.396 & 10.847 & 1.907 \\

& ReMoDiffuse \cite{112} & 3D & Enc-Dec & Hybrid Retrieval & \textcolor{blue}{\checkmark} & ICCV & 0.103 & 0.510 & 9.018 & 1.795 & 0.155 & 0.427 & 10.800 & 1.239 \\
\midrule

\multirow{5}{*}{\textbf{\parbox{1.5cm}{Unified \\ Model}}}
& Unify Motion \cite{130a} & 3D Rot. & Enc-Dec & Multi-Part Generation & \textcolor{orange}{\checkmark} & ArXiv & 1.167 & 0.548 & 14.720 & 2.010 & - & - & - & - \\
& Unify MoGPT \cite{123} & 3D Rot. & Enc-Dec &  Body-Controlled Gener. & \textcolor{red}{\xmark} & ArXiv & 0.191 & 0.496 & 9.860 & 2.137 & 0.614 & 0.427 & 11.256 & 2.357 \\
& MotionLLaMA \cite{115} & 3D Rot. & Autoregressive & Synthesis and Comprehen. & \textcolor{orange}{\checkmark} & ArXiv & 0.1133 & 0.4926 & 20.343 & 1.0637 & - & - & - & - \\
& MotionLLM \cite{52} & 3D Rot. & Autoregressive & Single and Multi  Gener. & \textcolor{orange}{\checkmark} & ArXiv & 0.045 & 0.482 & 9.838 & 3.138 & 0.781 & 0.409 & 11.410 & 2.982 \\
& UDE Engine \cite{148} & 3D Rot. &  Enc-Dec & Motion Refinement & \textcolor{red}{\xmark} & ECCV & 2.670 & 8.210 & 6.750 & 2.340 & - & - & - & - \\
\bottomrule
\end{tabular}
}
\end{table}

Distinct modes are identified based on clustering in the latent space or feature embeddings of the generated motions. Higher MMS indicates that the generative model effectively captures the diversity of motion categories, which is crucial for applications that require varied, human-like motions. The detailed metrics comparison along with their interpretation is shown in Table~\ref{tab:5}.

\subsubsection{Accuracy and Consistency Metrics}  
These metrics assess the degree to which the generated human motion aligns with conditional inputs, including text descriptions, audio, or other modalities. They are essential for assessing the quality of text-to-motion generation and ensuring that generated motions are both accurate and consistent with the input conditions.  

\textbf{Condition Consistency (Text-Motion Accuracy).}  
Condition consistency measures the accuracy with which the generated motions match the corresponding textual descriptions. For example, if a text prompt describes ``a person walking slowly'', the generated motion should visually and dynamically resemble slow walking. This metric is often computed by comparing the embeddings of the text and motion in a shared feature space, such as those learned by a multimodal model \cite{13,14,67,69,72,84,85,89,92,96,98,101,104,105,107,110,113,114,122,138}
. Higher consistency scores indicate that the model is effectively understanding and translating textual inputs into motion.  

\textbf{Text-Motion Consistency.} Text-motion consistency evaluates the coherence between the generated motions and their corresponding textual descriptions over the entire sequence. While condition consistency focuses on matching the input description to generated motions, text-motion consistency ensures that the generated motions remain semantically aligned with the text throughout the motion sequence. This can be assessed by calculating a similarity score between text and motion embeddings \cite{13,122,140,141,142,143,144,145,101,146,147,148,149,150,151,152}, ensuring the generated motion reflects the text in a temporally consistent manner. 
\textbf{R-Precision.}  
It quantifies the ranking accuracy of retrieved motion samples against a text query \cite{112}. It measures whether the generated motion is ranked among the top \( R \) relevant motions for a given textual prompt in a dataset. For example, if \(R=5 \), R-Precision checks if the correct motion appears in the top 5 most relevant results. Mathematically, it can be expressed as:  
\[
\text{R-Precision} = \frac{\text{Relevant motions in top-} R}{R}. \tag{\textcolor{blue}{19}}
\]

R-Precision is especially useful in retrieval-based evaluation setups, where the goal is to measure the alignment between the textual query and the retrieved motion samples. Higher R-Precision indicates better alignment and ranking accuracy \cite{13}, indicating that the model generates or retrieves motions most relevant to the input query.

\textbf{Motion Quality Metrics.}  
These metrics evaluate the realism, smoothness, and plausibility of the generated human motion sequences. They focus on ensuring that the generated motions not only match the intended task or context but also appear natural and fluid, closely resembling human-like movements.  

\textbf{Motion Quality.}  
Motion quality encompasses both subjective and objective evaluations of the naturalness and fluidity of generated motion sequences. Subjective evaluations rely on human judges rating the plausibility \cite{154} of the motion, whereas objective evaluations analyze specific motion attributes, such as velocity, acceleration, and joint coherence \cite{121}. High-quality motions exhibit smooth transitions, realistic joint movements, and adherence to biomechanical principles.

\textbf{Mean Per Joint Position Error (MPJPE).} It is an objective metric that quantifies the average error between the joint positions of the generated and ground-truth motions across all frames. It measures how closely the generated motion replicates the true joint positions in space, indicating motion accuracy. The formula for MPJPE is:

\begin{equation}
\text{MPJPE} = \frac{1}{T} \sum_{t=1}^{T} \frac{1}{J} \sum_{j=1}^{J} \left\| p_{t,j}^{\text{gen}} - p_{t,j}^{\text{gt}} \right\|_2 \tag{\textcolor{blue}{20}}
\end{equation}

where $T$ represents the number of frames in the motion sequence, and $J$ denotes the number of joints in the skeleton. The term $p_{t,j}^{\text{gen}}$ refers to the position of the $j$-th joint at frame $t$ in the generated motion, while $p_{t,j}^{\text{gt}}$ denotes the position of the $j$-th joint at frame $t$ in the ground-truth motion.

\begin{figure}[t]
    \centering
    \includegraphics[width=0.7\linewidth]{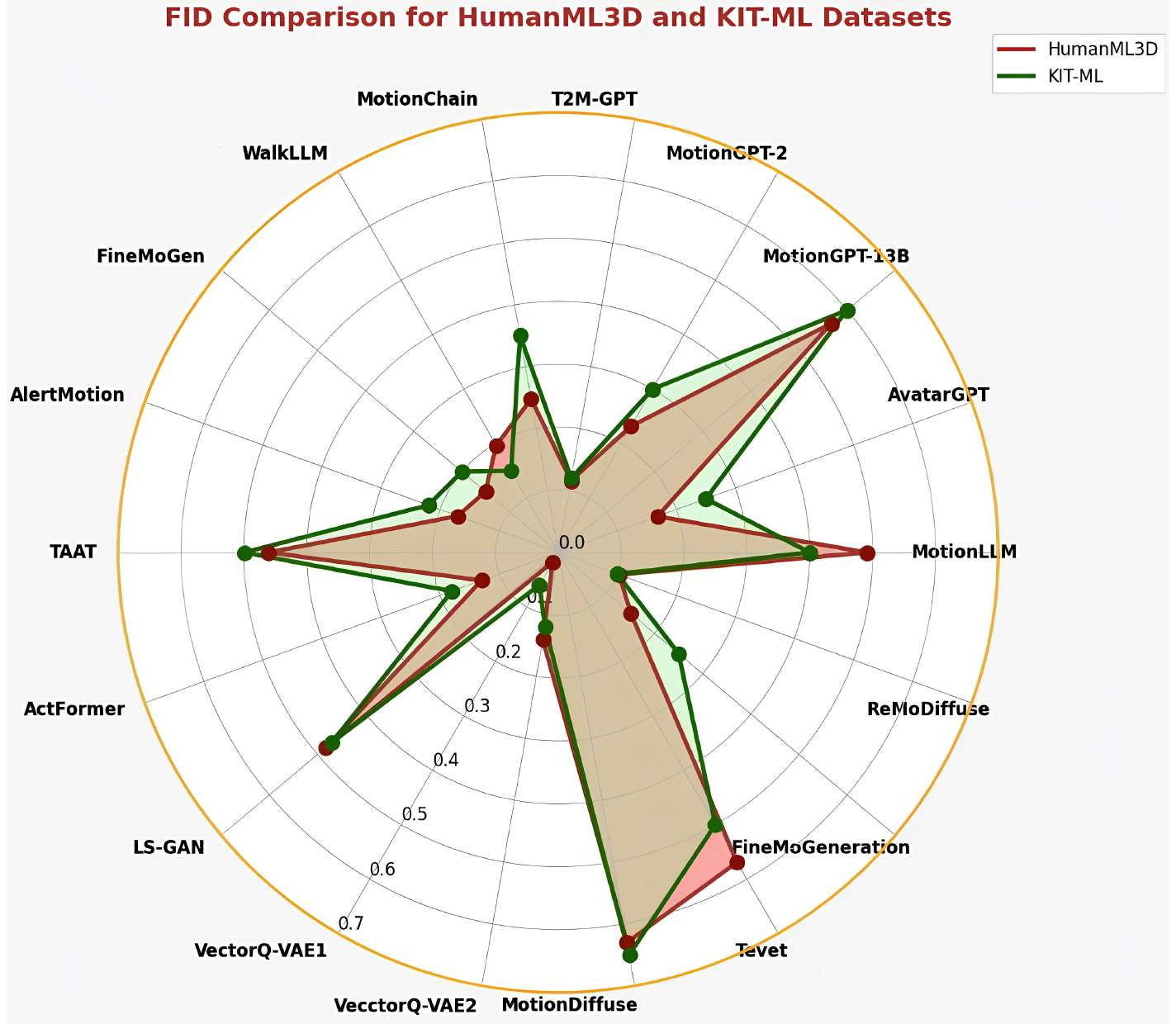}
    \caption{A comparative analysis of methods based on \textbf{FID} for Humanml3D and KIT-ML datasets.}
    \label{FIG:10}
\end{figure}

\vspace{1em} 

A lower MPJPE value indicates higher fidelity \cite{88,99,118,109,95,119} and closer alignment \cite{105} between the generated motion and the ground-truth motion.  
\begin{figure}[t]
    \centering
    \includegraphics[width=1\linewidth]{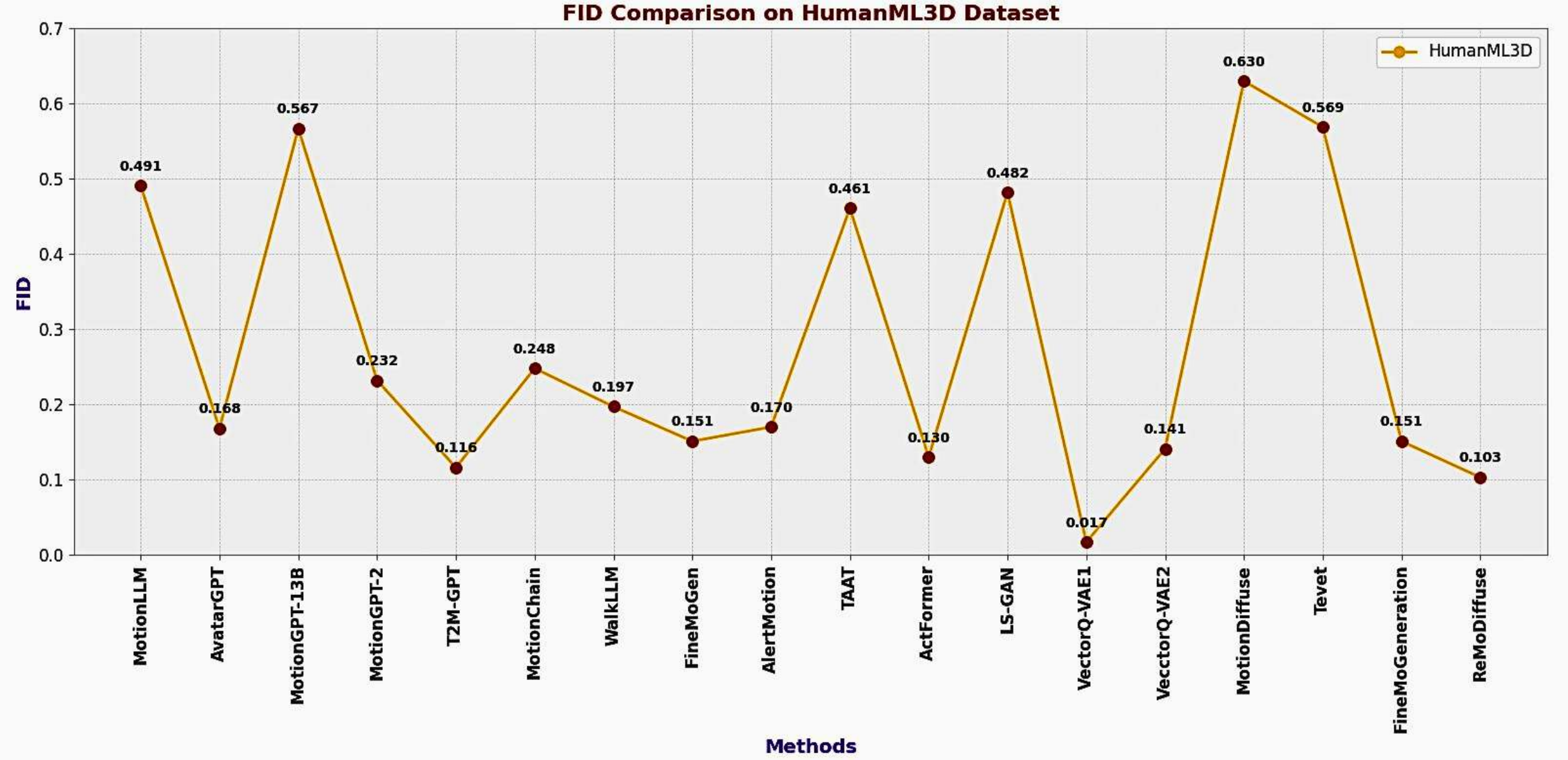}
    \caption{A comparative analysis of different methodologies based on the \textbf{FID} metric on the HUMANML3D dataset.}
    \label{FIG:11}
\end{figure}

\textbf{Importance of Unified Evaluation Metric in Human Motion Generation.} Motion quality metrics are crucial for benchmarking and comparing models effectively; however, the lack of standardization in evaluation methods poses challenges to reproducibility and fair assessment. Recent studies, such as \cite{22z}, emphasize the need for unified evaluation metrics to ensure consistency and comparability across diverse tasks, datasets, and motion representations. 
Similarly, \cite{156} advocates for metrics that align closely with human judgments to better evaluate the holistic quality of generated human motion. Table~\ref{tab:6}, Figure~\ref{FIG:10} and Figure~\ref{FIG:11}  illustrate the comprehensive evaluation metric based on FID results from recent publications of prominent methodologies on the two well-known HumanML3D and Kit-ML datasets.

\section{Potential Applications}

In recent years, advancements in \textcolor{blue}{GenAI}, particularly \textcolor{blue}{Multimodal LLMs} and \textcolor{blue}{Diffusion Models}, have transformed the landscape of \textbf{HMUG}. These cutting-edge technologies, with the emergence of the metaverse, have become integral to creating realistic and dynamic virtual humans in meta space, seamlessly integrating motion, text, and other modalities. Their impact is most notable in the realms of augmented reality (\textbf{AR}), virtual reality (\textbf{VR}), and the \textbf{metaverse}, where they enable immersive, interactive digital experiences. The scope of applications extends beyond entertainment into domains like healthcare, education, and training simulations, underscoring the transformative potential of multimodal human motion generation. Leading technology companies, including \textbf{Nvidia, Meta, Google, Microsoft, Samsung, Epic Games, Tencent, and Digital Domain} are spearheading initiatives such as virtual avatars, digital humans, and ``MetaHuman'' platforms, further accelerating innovation in this field. Below are key application domains where multimodal human motion generation could play a transformative role:

\subsection{Healthcare and Rehabilitation }

Therapeutic exercises can be enhanced by automatically generating tailored human motion sequences for physiotherapy and rehabilitation \cite{157,158,159,160}, addressing individual patient needs. Virtual simulations for gait analysis and correction provide a means to assess gait impairments and propose targeted exercises, facilitating personalized treatment \cite{161,162,163,164,165,165a}. Additionally, \textcolor{blue}{AI}-driven assistive technologies are being developed to support patients with motor disabilities by providing real-time motion predictions and training, further improving rehabilitation and daily functioning \cite{166,167}.

\subsection{ Virtual Reality (VR) and Augmented Reality (AR)} 

VR and AR applications are significantly enhanced through the generation of realistic human movements that adapt seamlessly to user interactions, creating immersive experiences \cite{168,169,170}. In gaming, this advancement enables non-player characters (NPCs) to exhibit more natural movements, thereby increasing player engagement \cite{171,173}. Furthermore, realistic human motion simulations are utilized in training scenarios across various professional domains, such as medicine, sports, and military operations, to provide effective and practical skill development \cite{173}.

\subsection{Entertainment and Fashion Industry}

In the entertainment and fashion industries, \textcolor{blue}{(AI}-powered human motion generation has revolutionized the integration of creativity and realism into digital content. In film and animation production \cite{174}, \textcolor{blue}{AI} automates the generation of realistic human movements, minimizing reliance on manual animation techniques and traditional motion capture, while enabling the creation of lifelike characters and immersive storylines. Game development benefits from diverse motion styles \cite {175,176,176a}, which enhance realism and engagement in gaming experiences. Similarly, in fashion \cite{177,178,180}, virtual models driven by \textcolor{blue}{AI}-generated motions bring dynamic, runway-like presentations to virtual try-on applications \cite{181,182} and digital fashion showcases, offering consumers an interactive and personalized experience that redefine retail and marketing strategies.

\subsection{Robotics and Human-Robot Interaction (HRI)}

Robotics and Human-Robot Interaction (HRI) significantly benefit from advancements in multimodal human motion generation. By learning from human motion patterns, these technologies enhance robotic motion planning and execution, enabling humanoid robots to perform tasks more efficiently and naturally \cite{183,184,185,186,187}. Additionally, they facilitate seamless collaboration between humans and robots by predicting and generating human-like movements, fostering more intuitive and effective interactions in both industrial and personal settings.

\subsection{Security and Surveillance}

Advancements in human motion generation play a critical role in enhancing security and surveillance systems. By analyzing and simulating human motion, these technologies enable the accurate prediction of behavior for detecting abnormal activities in real-time \cite{188,189}. Additionally, they facilitate forensic analysis by reconstructing human motions \cite{190,191,192} from available evidence, aiding in investigations and improving the overall efficacy of security protocols.

\subsection{Autonomous Vehicles} Human motion generation is integral to improving the safety and efficiency of autonomous vehicles by enabling accurate prediction and simulation of pedestrian movements. This technology enables vehicles to anticipate human behavior in complex scenarios, resulting in improved decision-making and enhanced collision avoidance. For instance, generative models have been used to simulate diverse pedestrian trajectories, enhancing the training of autonomous systems in real-world environments \cite{193,194,195}. These advancements underscore the importance of human motion understanding in developing reliable and adaptive autonomous driving solutions.

Multimodal human motion generation is a transformative technology with applications spanning healthcare, entertainment, robotics, education, and beyond. By integrating advancements in machine learning, \textcolor{blue}{GenAI}, and multimodal processing, this technology is paving the way for innovative solutions to complex challenges across industries. As the field advances, its potential to transform our interactions with digital and physical environments will continue to grow.

\section{Challenges and Future Work}

Despite the remarkable advancements in multimodal \textcolor{blue}{GenAI} for motion understanding and generation, noteworthy struggles remain in developing more efficient, scalable, and coherent models across diverse modalities. Multimodal diffusion models, autoregressive and LLM architectures, and transformer-based unified frameworks each have their strengths and places where improvement needs to be done, that face unique limitations, still challenging in their real-world applicability. These include computational inefficiencies, modality-specific constraints, and challenges in modeling long-range dependencies. Addressing these areas is crucial for enabling efficient and high-quality human motion generation, seamless cross-modal alignment, and real-time applications. In this section, we discuss the key areas of research in \textbf{HMUG} that aim to improve these architectures, benchmarks, and evaluation metrics. We outline promising future directions to guide further research in multimodal models.

\textbf{
Multimodal Architecture Level Challenges.} Multimodal LLMs with an autoregressive probabilistic nature as shown in Figure~\ref{FIG:9} have become dominant in visual understanding and generation tasks. However, they face challenges in visual tokenization, representation learning, and adaptability across tasks. Existing tokenization techniques, such as VQGAN \cite{83}, struggle to handle large codebooks, resulting in inefficiencies in multi-scale data compression. Some improvements have been made through VAE variants, such as VQ-VAE \cite{82,92,93} and the most recent RVQ-VAE \cite{17,100,101}, as well as VQ-diffuse \cite{121}, but limitations persist. The choice between discrete and continuous representations presents another trade-off: while continuous representations allow for smoother transitions, they introduce training instabilities. Additionally, the inductive biases of purely autoregressive architectures make them less effective for visual data, requiring hybrid approaches that incorporate structured priors and hierarchical tokenization \cite{6,8,118,122}. Unlike \textcolor{blue}{MLLMs}, visual autoregressive models lack generalization to downstream tasks, limiting their adaptability. Moreover, modeling long-range dependencies in human motion video remains challenging, necessitating more efficient causal attention mechanisms. Early fusion approaches, which integrate multiple modalities at the input level, are computationally exhaustive for motion video generation, further complicating real-time applications. Alternative alignment architectures, as shown in Figure~\ref{FIG:5} and Figure~\ref{FIG:8}, offer potential improvements in efficiency and effectiveness but require further refinement.
Future advancements should focus on enhancing hierarchical visual tokenization, developing improved loss functions for continuous modeling, integrating structured priors, and designing task-adaptive pre-trained models. These improvements will enhance performance, generalization, and efficiency in multimodal autoregressive models for human motion generation.

Multimodal diffusion architectures \cite{22}, as shown in Figure~\ref{FIG:7} and Figure~\ref{FIG:9}, achieve remarkable improvements in human motion generation while mitigating network collapse issues. However, they face significant challenges related to computational efficiency, scalability, and modality integration. The iterative denoising process incurs high computational costs and inference latency, rendering real-time applications challenging. Long-sequence motion generation remains constrained by high training costs and complex spatiotemporal dependencies. Additionally, integrating diverse modalities such as text, vision, audio, and motion is challenging due to their structural differences, while ensuring realism and plausibility in generated outputs remains an ongoing concern.
Future research should focus on optimizing latent-space diffusion \cite{9, 121b} to reduce time complexity and explore alternative generative models, such as GANs, which require a single forward pass for faster processing. While these models currently lag behind diffusion-based approaches, their improved diversity and applicability in multimodal tasks could enhance long-motion content generation. Unifying the strengths of diffusion, GAN, and VAE architectures, as shown in Figure~\ref{FIG:7}, will be essential for advancing multimodal diffusion models. This involves improving hierarchical tokenization, integrating physics-informed priors, and refining progressive generation strategies. Additionally, optimizations such as model pruning, distillation, and approximate inference are crucial for enabling real-time applications. 

A unified framework \cite{115,123,148,22z} offers significant advantages by leveraging the complementary strengths of both architectures as depicted in Figure~\ref{FIG:9}. Multimodal LLMs excel at understanding and generating structured sequences across text and motion, while diffusion models generate high-quality outputs with superior diversity and realism. Ensuring synchronization across different modalities, such as motion-text correspondence, remains difficult due to inconsistencies in data representations. Some works \cite{119,121,120} employ the connector strategy, utilizing a dense or a mixture of expert models to address the synchronization issues illustrated in Figure~\ref{FIG:8}. The advanced strategy involves training both models in a shared high-dimensional embedding space. Large-scale multimodal transformers, as shown in Figure~\ref{FIG:9c}c, help cooperate both models in alignment and shared space, enhancing comprehension and generation capabilities \cite{30,100,151}. However, also suffer from gradient instability and mode collapse, making training optimization a key research priority. The high computational cost of transformer-based models further limits their real-world applicability, as inference remains slow and resource-intensive. Generating long-form video and motion sequences is another significant challenge, as most models are limited to short-term outputs. Future research should focus on refining the above strategies and cross-attention mechanisms, as well as adaptive training strategies such as reinforcement learning, and developing lightweight architectures to reduce inference costs. Dynamic conditioning methods, such as prompt-based autoregressive models, could enhance long-term coherence, while real-time applications would benefit from efficient parallel computing and edge \textcolor{blue}{AI} solutions.

\textbf{Multimodal Datasets and Benchmarks.}
The development of multimodal datasets and benchmarks for human motion generation faces several significant challenges. One of the main issues is the lack of large-scale, unified datasets that can effectively support cross-validation \cite{13} and evaluation across different tasks. Current datasets often focus on either understanding or generation tasks, but do not integrate both aspects in a unified manner. This fragmented approach hinders the comprehensive assessment of models that simultaneously handle multiple tasks \cite{124,125,126}. Additionally, data scarcity, particularly in human motion video generation, remains a significant challenge due to privacy concerns, Erroneous data quality with inconsistent human annotations and descriptions, and the high costs associated with data collection. Future work should focus on creating diverse, high-quality datasets that not only cover a wide range of human activities but also include rich annotations across multiple modalities. Furthermore, designing unified benchmark datasets that combine both understanding and generation tasks is critical for more accurate model evaluation, diversity, and improvement in this domain. Expanding data sources and addressing privacy concerns will be key to developing robust models that can generalize well across various applications.

\textbf{Evaluation Metrics.}
The authors have made commendable progress in developing evaluation metrics, contributing valuable insights to the field. Their work has laid a solid foundation for assessing models across various domains \cite{30,100,123}. However, as the field continues to evolve, there is a clear and pressing need for a unified evaluation framework. While the proposed metrics are certainly a step in the right direction, there is room for further refinement and expansion. The establishment of more appropriate evaluation metrics will not only improve the accuracy and reliability of assessments but also address gaps in current practices. A unified evaluation metric \cite{22z}, which combines diverse metrics into a cohesive structure, is urgently needed. This approach will provide a consistent and comprehensive method of evaluation, ensuring that models and methods are assessed in a way that is both fair and transparent across the board.
The need for such a unified framework is clear, and the authors' contributions thus far pave the way for future advancements in this area.

\textbf{Applications. }
Human motion applications, particularly in the realm of lightweight multimodal \textcolor{blue}{GenAI}, face several challenges that require innovative solutions. One major hurdle is the lack of photorealism, with a need for more natural, lifelike human motion \cite{174,178}, including accurate hand and body movements. Additionally, expanding the duration and refining the control of motion \cite{166} generation—from short clips to longer, more dynamic videos with multi-person interactions—remains a significant challenge. Future advancements should prioritize improving the generation of long-duration human videos while also enhancing the semantic depth and interactivity of motion. Another critical direction is the integration of these models with emerging technologies, such as AR/VR \cite{168,170} and robotics \cite{169a, 193,194,195}, which enables more immersive and interactive experiences. Furthermore, as \textcolor{blue}{AI} continues to evolve, there is a need to adapt multimodal \textcolor{blue}{GenAI} systems \cite {22} to function effectively in dynamic environments, ensuring that human motion generation remains efficient, controllable, and context-aware. These advancements have broad applicability across diverse fields, including fashion, health, autonomous vehicles, and education, where lifelike, controlled motion generation can enhance user experiences, improve training, and offer innovative solutions.

\textbf{Ethical and Adversarial Effects.}
Ethics and adversarial impacts pose significant challenges \cite{72} in the development and deployment of motion models, especially with the rise of digital humanoids and virtual humans \cite{73,76,77,78}. Ensuring fairness and ethical considerations is paramount, particularly regarding privacy and the responsible use of human-generated data. Providing privacy-free human image data can help alleviate concerns surrounding personal data protection, but it also raises issues related to consent and the potential for misuse. Adversarial effects, such as the manipulation or generation of misleading or harmful content \cite{69}, remain a significant concern, particularly when models are applied in sensitive domains like healthcare, education, or entertainment.

To address these challenges, future work should focus on creating robust ethical frameworks that govern the creation, use, and distribution of motion models. These frameworks should prioritize fairness, transparency, and accountability, while also ensuring that the generated content does not perpetuate bias or harm. Additionally, advancements in adversarial training and model robustness are needed to prevent malicious exploitation of these technologies. Moving forward, the field must strike a balance between innovation and ethical responsibility, ensuring that generative models are developed and used in ways that benefit society while minimizing potential risks.

\section{Conclusion}
In this review paper, we explored recent advancements in data representation, \textcolor{blue}{MLLMs}, autoregressive models, multimodal diffusion models, and emerging unified frameworks that integrate these technologies. We also examined the role of multimodal transformers in enhancing \textbf{HMUG}, as well as their broader applications across diverse fields. By providing a comprehensive overview of the architectures and key contributions in this area, we aim to offer valuable insights into the current landscape and highlight opportunities for future research.

Despite significant progress, several challenges persist, including improving photorealism, extending motion durations, refining control mechanisms, developing unified evaluation metrics, and addressing ethical concerns. Future work should focus on addressing these issues, enhancing model scalability and robustness, and exploring integration with emerging technologies, such as augmented and virtual reality (AR/VR) and robotics. These advancements have the potential to transform industries such as fashion, healthcare, autonomous systems, and education, enabling more interactive, efficient, and context-aware applications.



\bibliographystyle{cas-model2-names}

\begin{thebibliography}{00}
\bibitem{4} Jong Wook Kim, Alex Nichol, Yang Song, Lijuan Wang, Tao Xu, “DALL·E 3 understands significantly more nuance and detail than our previous systems, allowing you to easily translate your ideas into exceptionally accurate images.”
\end{thebibliography}

\end{document}